\documentclass[9pt,twocolumn,twoside]{pnas-new}

\articletype{CLASSIFICATION}

\templatetype{pnasresearcharticle} 

\usepackage{algorithm}
\usepackage{algpseudocode}
\usepackage{adjustbox}
\usepackage{multirow}
\usepackage{subcaption}

\begin{document}

\title{Neural operator-based digital twins for modeling amyloid-$\beta$ and tau propagation and treatment optimization in Alzheimer's disease}



\author[a]{Xiaofeng Xu}

\author[b]{Tingting Dan}

\author[c]{Zifan Zhou}

\author[c]{Bin Li}

\author[b,d,e,f]{Guorong Wu}

\author[a]{Wenrui Hao}

\affil[a]{Department of Mathematics, Pennsylvania State University, University Park, PA 16802, USA}

\affil[b]{Department of Psychiatry, University of North Carolina at Chapel Hill, Chapel Hill, NC 27599, USA}

\affil[c]{School of Electrical Engineering and Computer Science, Pennsylvania State University, University Park, PA 16802, USA}

\affil[d]{Department of Computer Science, University of North Carolina at Chapel Hill, Chapel Hill, NC 27599, USA}

\affil[e]{Department of Statistics and Operation Research, University of North Carolina at Chapel Hill, Chapel Hill, NC 27599, USA}

\affil[f]{UNC Neuroscience Center, University of North Carolina at Chapel Hill, Chapel Hill, NC 27599, USA}

\leadauthor{Xu}

\significancestatement{
Alzheimer’s disease progression exhibits substantial inter-individual variability, making accurate prediction and personalized intervention particularly challenging.
We develop a brain digital twin framework that combines reaction--diffusion modeling with neural operator learning to infer patient-specific disease dynamics directly from longitudinal neuroimaging data.
The resulting digital twin accurately reconstructs and predicts the propagation of amyloid-$\beta$ and tau pathology, and enables individualized treatment optimization through optimal control.
An immersive virtual reality platform further supports interactive visualization and exploration of disease trajectories.
This work demonstrates how data-driven scientific computing and mechanistic modeling can be combined to yield interpretable, patient-specific representations of neurodegenerative disease progression, providing a foundation for personalized prediction and intervention.
}


\authordeclaration{The authors declare no competing interest.}

\correspondingauthor{To whom correspondence should be addressed. E-mail: wxh64@psu.edu}
\authorcontributions{Please provide details of author contributions here.}



\begin{abstract}
Accurately predicting the spatiotemporal evolution of amyloid-$\beta$ and tau proteins at the individual level is critical for improving the diagnosis and treatment of Alzheimer’s disease. We consider the problem of constructing patient-specific digital twins that model the propagation of these biomarkers on the cortical surface using reaction--diffusion dynamics.
A major challenge is that the underlying nonlinear aggregation mechanisms are unknown and must be inferred from sparse, noisy, and heterogeneous longitudinal PET imaging data.
To address this, we develop a data-driven framework that learns biomarker dynamics directly from clinical observations.
The approach combines operator learning with reduced-order representations to infer governing equations of disease progression from data.
Using this framework, we achieve predictive accuracies of 87\% for amyloid-$\beta$ and 81\% for tau.
Building on the learned dynamics, we further formulate a PDE-constrained optimal control problem to design personalized therapeutic strategies that regulate pathological protein propagation. By integrating data-driven dynamical modeling with treatment optimization, the proposed digital twin framework provides an interpretable and predictive platform for understanding disease progression and enabling precision interventions in neurodegenerative disorders.
\end{abstract}

\dates{\today}
\doi{\url{www.pnas.org/cgi/doi/10.1073/pnas.XXXXXXXXXX}}

\maketitle
\thispagestyle{firststyle}
\ifthenelse{\boolean{shortarticle}}{\ifthenelse{\boolean{singlecolumn}}{\abscontentformatted}{\abscontent}}{}

\Firstpage





Alzheimer’s disease (AD) is characterized by the progressive accumulation and spatial propagation of amyloid-$\beta$ and tau proteins, whose distributions serve as key biomarkers of disease onset and progression. Understanding and predicting the evolution of these biomarkers is essential for early diagnosis and for developing personalized therapeutic strategies. Recent advances in longitudinal neuroimaging, particularly through the Alzheimer's Disease Neuroimaging Initiative (ADNI), have generated surface-based measurements of amyloid-$\beta$ and tau over multiple time points, providing unprecedented opportunities to study their spatiotemporal dynamics on the cortical manifold and to construct patient-specific digital twins of disease progression.

Mathematical models based on reaction--diffusion partial differential equations (PDEs) provide a natural framework for describing the spread of pathological proteins in the brain, with diffusion terms representing spatial propagation and reaction terms accounting for local growth, decay, and aggregation processes \cite{rabiei2025data,petrella2019computational,zheng2022data,petrella2024personalized,wang2026learning,li2025data}. 
Existing computational models have incorporated important biological insights into amyloid-$\beta$ and tau pathology, but the underlying nonlinear reaction mechanisms remain incompletely understood\cite{raj2012network,raj2015network,yu2023uncovering, zhang2024discovering}. A major challenge is that these mechanisms must be inferred from sparse, noisy, and heterogeneous longitudinal observations obtained in clinical studies.

Recent developments in scientific machine learning \cite{raissi2019physics,karniadakis2021physics}, have created new opportunities for integrating physical principles with data-driven models to identify governing equations directly from observations. In particular, operator learning methods, including DeepONet\cite{lu2021learning} and Fourier Neural Operators\cite{li2020fourier}, have shown remarkable capability in approximating nonlinear mappings arising from complex dynamical systems. More recently, the Laplacian Eigenfunction Neural Operator (LENO) has been introduced to learn nonlinear reaction operators in reaction--diffusion systems while naturally accommodating geometries\Parasplit represented by manifolds. These advances suggest the possibility of constructing digital twins that combine mechanistic disease models with patient-specific longitudinal data to predict future disease trajectories.

\begin{figure*}[h] \centering \includegraphics[scale=0.36]{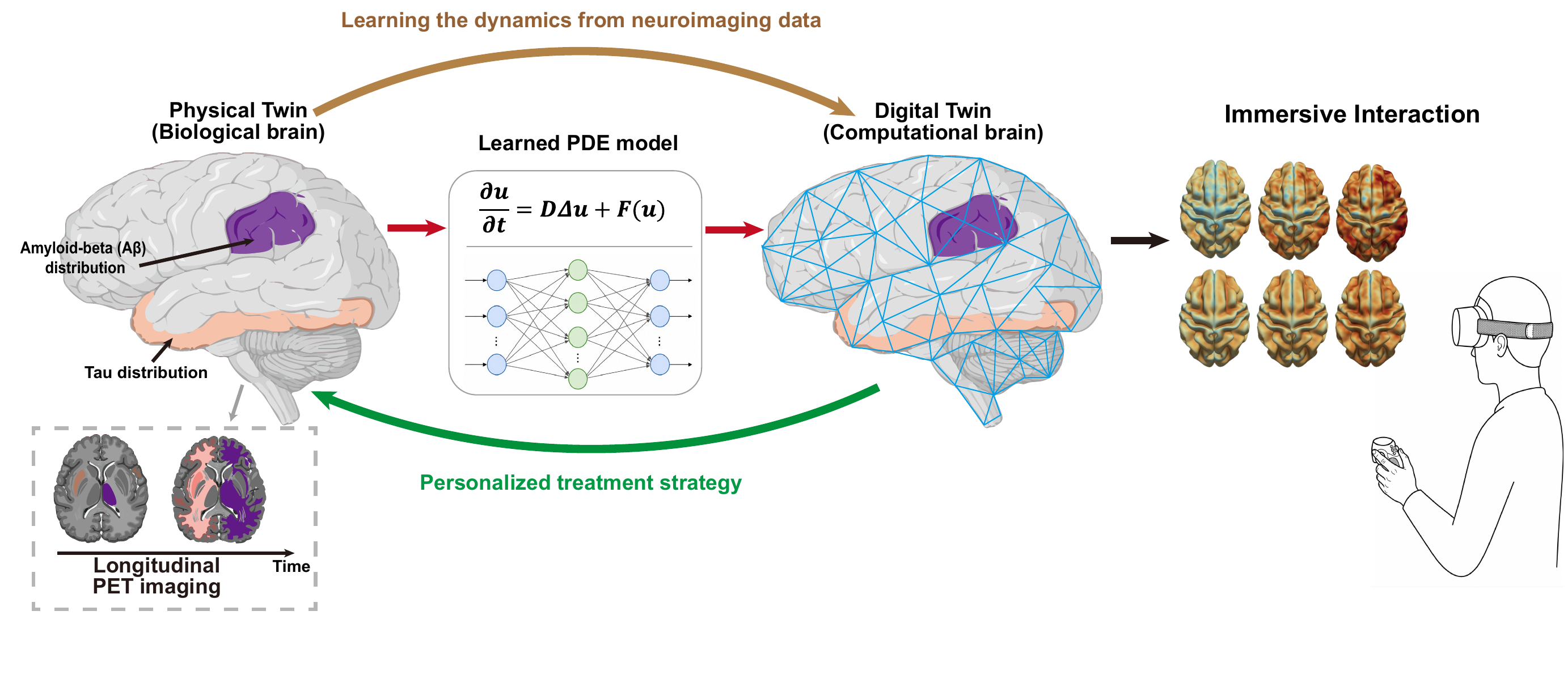} \caption{ \textbf{Digital twin framework for personalized modeling and treatment of Alzheimer's disease.}  Longitudinal neuroimaging measurements of amyloid-$\beta$ (A$\beta$) and tau on the cortical surface are integrated with a reaction--diffusion model to construct a patient-specific digital twin. In the forward mode, the framework learns biomarker dynamics from sparse longitudinal observations and predicts the spatiotemporal progression of pathology. In the backward mode, the learned digital twin is coupled with PDE-constrained optimal control to identify personalized therapeutic interventions that regulate disease evolution. The framework further enables immersive visualization and interactive exploration of disease trajectories and treatment outcomes through a virtual reality platform. } \label{fig:DTschematic} \end{figure*}

Here, we develop a digital twin framework that integrates longitudinal neuroimaging measurements with operator learning to infer biomarker dynamics on cortical surfaces. Using LENO, we learn nonlinear reaction terms governing amyloid-$\beta$ and tau propagation directly from sparse and irregular observations. The learned models are then coupled with a PDE-constrained optimal control framework to determine individualized therapeutic interventions that regulate biomarker progression. As illustrated in Fig.~\ref{fig:DTschematic}, the framework operates in two complementary modes: a forward mode that predicts the spatiotemporal evolution of pathological proteins and a backward mode that identifies treatment strategies capable of slowing disease progression \cite{lee2025optimal,hao2022optimal}. To facilitate clinical interpretation and interaction, we further incorporate virtual reality visualization, enabling immersive exploration of biomarker propagation and treatment effects.

By unifying operator learning, mechanistic disease modeling, optimal control, and interactive visualization, this work establishes a computational framework for personalized digital twins of Alzheimer’s disease. The framework provides a data-driven approach for predicting biomarker evolution and designing patient-specific interventions, offering a foundation for precision medicine in neurodegenerative disorders.

\section{Results}


\subsection{Problem setup and model description}

Longitudinal neuroimaging data provide a noninvasive window into the spatiotemporal progression of Alzheimer's disease (AD) pathology. In this study, we use longitudinal amyloid-$\beta$ PET and tau PET measurements from the ADNI cohort, mapped onto cortical surface representations across multiple time points and subjects. This yields biomarker fields defined on a two-dimensional manifold embedded in $\mathbb{R}^3$.

Let $\Gamma \subset \mathbb{R}^3$ denote the cortical surface. We model the temporal evolution of a biomarker concentration $u(x,t)$ on $\Gamma$, where $x \in \Gamma$ and $t \ge 0$, using a surface reaction--diffusion equation
\begin{equation}\label{eq:model_strong_form}
\partial_t u(x,t) - D \Delta_{\Gamma} u(x,t) = \mathcal{F}(u(x,t)),
\end{equation}
where $D>0$ is a diffusion coefficient, $\Delta_{\Gamma}$ is the Laplace--Beltrami operator \cite{Atkinson2012}, and $\mathcal{F}$ is an unknown nonlinear reaction operator.

The diffusion term captures spatial propagation along the cortical manifold, while the reaction term encodes local production, aggregation, and clearance processes. Since $\mathcal{F}$ is unknown, it is inferred directly from longitudinal data through an operator learning framework.
Accordingly, the learning objective is to identify nonlinear reaction dynamics consistent with temporal transitions observed between consecutive scans across subjects.

Beyond modeling biomarker evolution, we extend the system to incorporate therapeutic intervention by introducing a control term representing treatment intensity. This yields a PDE-constrained optimal control formulation in which the learned dynamics are used to determine dosing strategies that modulate disease progression while balancing treatment cost.

Together, the learned reaction--diffusion model and the control formulation define a patient-specific digital twin that enables both forward simulation of biomarker evolution and inverse design of intervention strategies.
Beyond these predictive and prescriptive capabilities, we further incorporate an immersive virtual reality environment to facilitate intuitive exploration of disease progression and therapeutic response.


\subsection{Population-level digital twin}

Our population-level digital twin is trained using the LENO method described in Section~\ref{sec:materials-methods}, with amyloid-$\beta$ and tau data modeled separately for the left and right hemispheres.
The learned models achieve high predictive accuracy for amyloid-$\beta$ (above 87\%) and moderate predictive accuracy for tau (above 81\%), with consistent performance across hemispheres.
Model performance is evaluated using relative $L^2$ errors in both spectral and physical spaces (Acc$_1$ \% and Acc$_2$ \%, see Methods).
Training and test results for amyloid-$\beta$ and tau are summarized in Table~\ref{tab:population-acc}.
For all experiments, the nonlinear reaction operator is approximated using a shallow fully connected neural network with width $M$ and \texttt{tanh} activation.
We additionally conducted extensive hyperparameter searches over truncation levels, network depth, width, activation functions, and weight decay parameters.
Detailed results are provided in \textcolor{blue}{SI Appendix 2}. 

\begin{table}[H]
\centering
\setlength{\tabcolsep}{5pt}
\renewcommand{\arraystretch}{1.2} 
\begin{tabular}{|l|l|c|c|c|c|}
\hline
\multirow{2}{*}{Metrics} & \multirow{2}{*}{Phase} & \multicolumn{2}{c|}{ Amyloid-$\beta$} & \multicolumn{2}{c|}{ Tau} \\ \cline{3-6} 
 &  & Left & Right & Left & Right \\ \hline
\multirow{2}{*}{Acc$_1$ \%} 
 & Train & $90.8\pm7.0 $ & $89.2\pm 7.2$&  $84.0 \pm 8.2$& $83.0 \pm 8.5$\\ \cline{2-6} 
 & Test  & $87.7 \pm 3.9 $  & $88.2 \pm 3.6$& $82.5 \pm 2.3$&  $ 82.4 \pm 1.6$\\ \hline
\multirow{2}{*}{Acc$_2$ \%} 
 & Train & $89.8 \pm 6.7$ &  $88.4 \pm 7.0$  & $82.3 \pm 8.1$& $ 81.5\pm 8.3$\\ \cline{2-6} 
 & Test & $ 87.2 \pm 3.9$ & $87.7 \pm 3.6$&   $81.1 \pm 2.7$&$81.0 \pm 2.1 $ \\ \hline
\end{tabular}
\caption{
Train and test accuracy of the learned model.
Values are reported as mean $\pm$ standard deviation across subjects in the training cohort.
The network widths are $M = 1024$ and $M = 128$ for amyloid-$\beta$ and tau, respectively.  
The spectral truncation level is $P=4096$. 
}
\label{tab:population-acc}
\end{table}

\begin{figure*}[t!]
    \centering
    \includegraphics[width=0.9\linewidth]{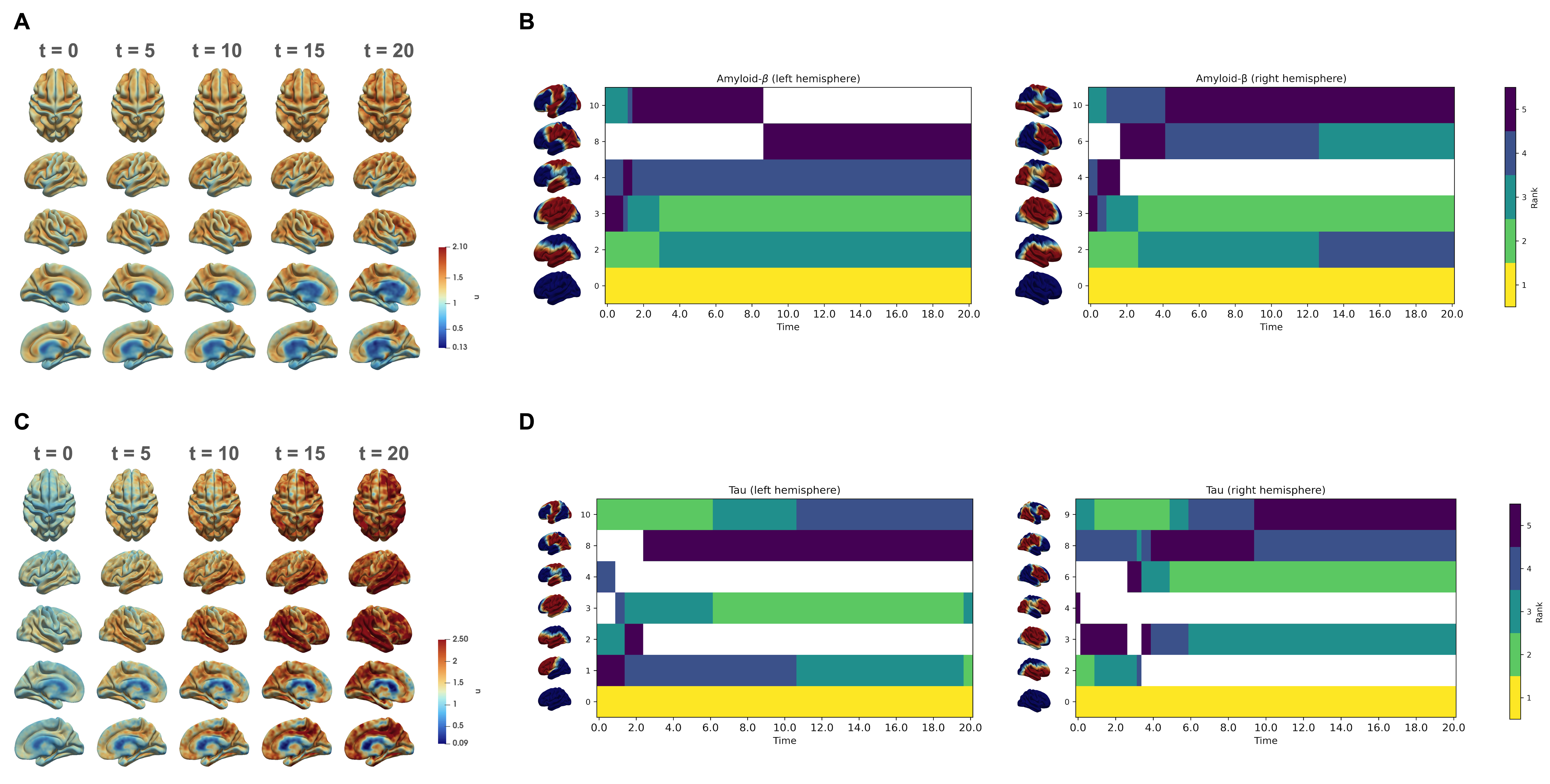}
\caption{
\textbf{Population-level digital twin simulations of amyloid-$\beta$ and tau propagation on the cortical surface and their spectral representations.}
(A) Simulated spatiotemporal evolution of amyloid-$\beta$ from the population-average initialization.
(B) Temporal evolution of the top five dominant Laplace--Beltrami eigenmodes for amyloid-$\beta$ in the left and right hemispheres. Colors indicate modal rank, with rank 1 corresponding to the most active mode.
(C) Same as (A) for tau.
(D) Same as (B) for tau.
}
\label{fig:population-digital-twin}
\end{figure*}

We now use the learned digital twin to simulate the evolution of amyloid-$\beta$ and tau on the brain cortical surface, starting from an initial state set to be the population average, computed from the earliest available measurement for each subject. 
The simulated results of amyloid-beta and tau at different time points are shown in Figure~\ref{fig:population-digital-twin}. 

The population-level simulations exhibit progressive spatiotemporal evolution of both amyloid-$\beta$ and tau on the cortical surface (Fig.~2A,C).
Starting from the population-average initialization, amyloid-$\beta$ shows gradual accumulation with relatively diffuse spatial patterns, whereas tau develops stronger spatial contrast and more region-specific amplification over time, indicating increased heterogeneity in its propagation dynamics. 
The corresponding top five ranked spectral modes (Fig.~2B,D) indicate that the dynamics are dominated by low-frequency Laplace--Beltrami eigenmodes, with their relative contributions evolving over time. 
Visual inspection suggests that the dominant modes reflect large-scale spatial patterns spanning medial frontal, parietal, and temporal association cortices.
While the dominant modes persist across the simulation, their rankings vary, reflecting a temporal reorganization of the principal spatial patterns rather than the emergence of high-frequency components. 
This low-dimensional structure suggests that the large-scale progression of both biomarkers is governed primarily by smooth global modes. 


\subsection{Patient-specific digital twin simulation and prediction for individuals}
To further examine how the learned dynamics manifest at the individual level, we apply the trained model to simulate the evolution of amyloid and tau for each subject.
To account for inter-subject variability, we adopt a transfer learning strategy by fine-tuning the pretrained model on subject-specific data, enabling efficient personalization while preserving the shared disease dynamics learned from the population.



\begin{figure*}[t!]
    \centering
    \includegraphics[width=0.9\linewidth]{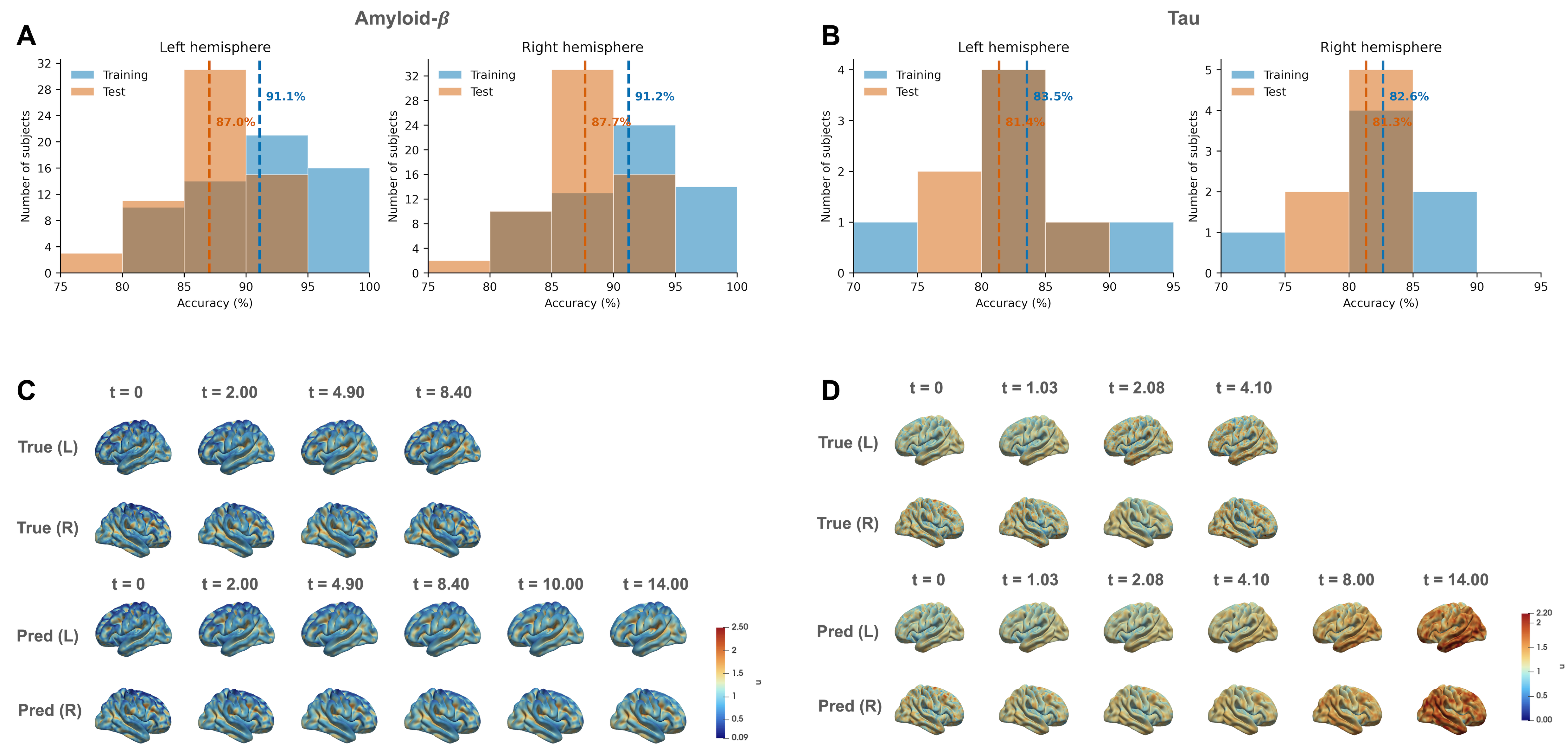}
\caption{
\textbf{Patient-level performance of the digital twin model for amyloid-$\beta$ and tau.}
(A--B) Distributions of training and test accuracies across subjects in the left and right hemispheres for amyloid-$\beta$ (A) and tau (B). Dashed lines indicate mean accuracies.
(C--D) Spatiotemporal evolution of cortical amyloid-$\beta$ (C) and tau (D) for a representative subject. The top two rows (true) show observed concentrations at available time points, where the earlier scans are used for training, and the final scan is reserved for testing.
The bottom two rows (pred) show corresponding model predictions over time.
Predictions at later time points extend beyond the final observed scan and therefore do not have corresponding ground-truth references.
}
\label{fig:patient_level_performance}
\end{figure*}

Figures~\ref{fig:patient_level_performance}A,B show the distributions of training and test accuracies across subjects for both hemispheres for amyloid-$\beta$ and tau, respectively. For amyloid-$\beta$, the model achieves high accuracy with a small generalization gap, indicating strong consistency of the learned dynamics across subjects. For tau, the model attains a mean test accuracy of approximately 81\%, which is lower than that of amyloid-$\beta$, likely reflecting the smaller sample size and higher noise level in the tau data. Nevertheless, in both cases, the relatively small shift between training and test distributions suggests stable generalization without significant overfitting.

Representative spatiotemporal evolutions for a single subject are shown in Fig.~\ref{fig:patient_level_performance}C,D for amyloid-$\beta$ and tau, respectively. The model captures the overall spatial patterns and temporal trends observed in the data, while providing consistent forward predictions beyond the observed time points.

\subsection{Optimal control enables treatment design}

We now introduce a treatment term into our learned disease dynamics to explore treatment strategies.
The treatment model is given by 
\begin{equation}\label{eq:treatment_model}
    \partial_t u - D \Delta_{\Gamma} u = \mathcal{F}(u) - C(t)u, \quad x \in \Gamma,
\end{equation}
where $\mathcal{F}(u)$ is now the learned nonlinear dynamics and $C(t)$ is the dosing intensity of a treatment.
Since we decouple the reaction dynamics of amyloid-$\beta$ and tau, the variable $u$ stands for either amyloid-$\beta$ or tau. 
We seek an optimal dosing function $C(t)$ that minimizes
\begin{equation}\label{eq:cost}
J(u,C)
=
\int_0^T \left( \int_{\Gamma} u(x,t)^2 dx + \alpha(t) C(t)^2\right)\,dt, 
\end{equation}
subject to the dynamics \eqref{eq:treatment_model}. 
Here, $\alpha(t)>0$ is a time-dependent penalty weight that penalizes treatment intensity, reflecting the side effects of medication or treatment.
It regulates the trade-off between minimizing plaque burden and limiting the side effects of anti-plaque treatment \cite{van2023lecanemab, sims2023donanemab}.

The mathematical details on solving the optimal control problem can be found in Section \ref{sec:appendix-optimal-control} and \textcolor{blue}{SI Appendix 6}.

\subsection{Population-level optimal control}\label{subsec:control-numeric-comparison}
In the population-level optimal control experiments, we consider two distinct treatment side-effect mechanisms represented through different penalty weights $\alpha(t)$.
The first employs a constant penalty, corresponding to treatment side effects that remain uniform throughout the treatment period.
The second uses a time-dependent penalty that decays over time, reflecting stronger side effects during the early stages of treatment that gradually diminish as patients adapt to the treatment.
The initial condition is chosen as the population-average biomarker distribution.
To solve the resulting optimality system from the optimal control problem, we employ the forward--backward sweep algorithm described in \textcolor{blue}{SI Appendix 6}.

\begin{figure*}[h]
    \centering
    \includegraphics[width=0.6\linewidth]{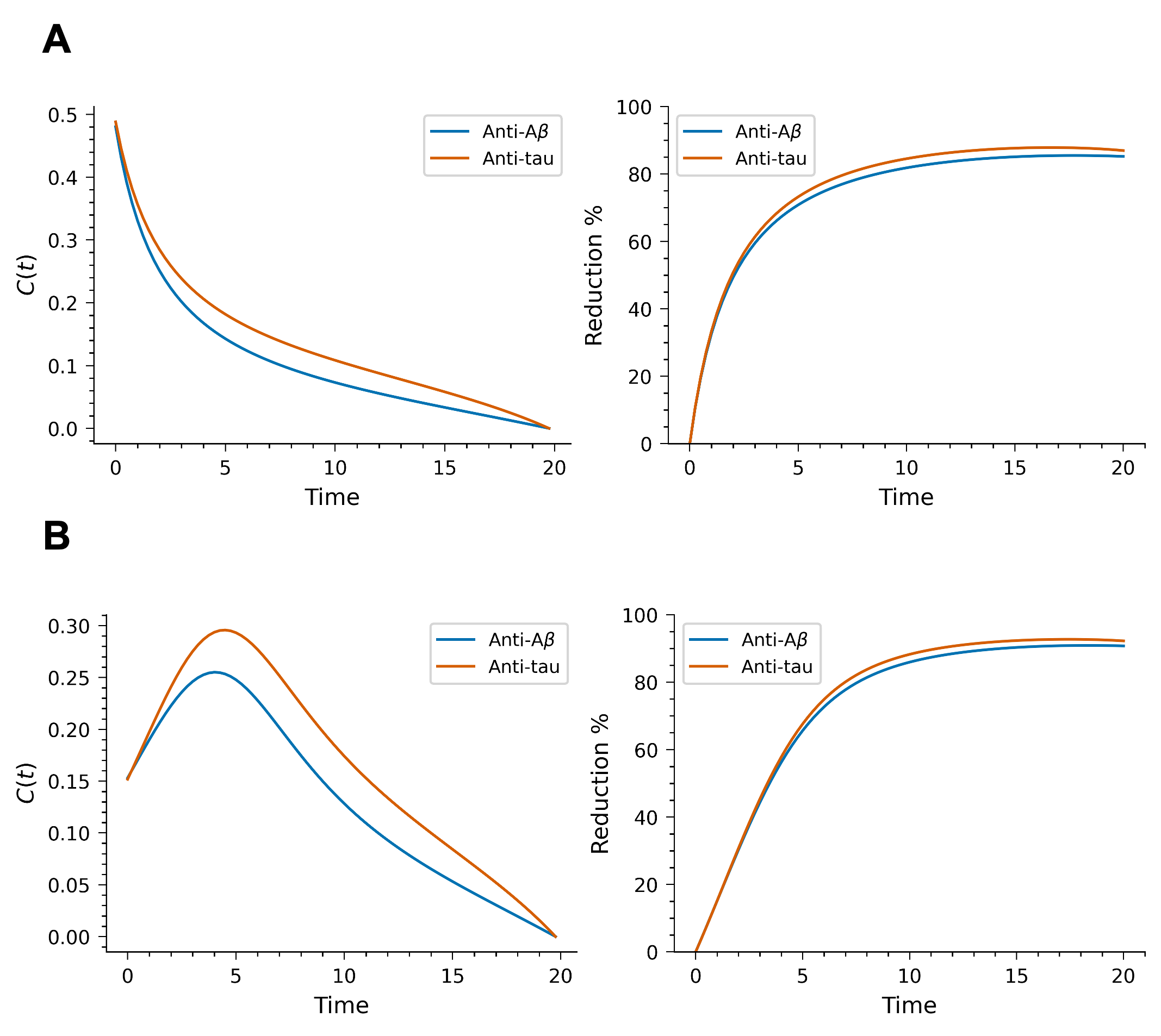}
\caption{
{\bf Comparison of optimal dosing strategies and treatment outcomes under constant and time-dependent penalty weights for amyloid-$\beta$ and tau.} 
(A) Constant penalty $\alpha=4\times10^5$. Left: optimal control $C(t)$ for anti-amyloid-$\beta$ and anti-tau treatment. 
Right: evolution of the reduction percentage of the mean amyloid-$\beta$ and tau concentration over time. 
(B) Time-dependent penalty $\alpha(t)=\alpha_1+\alpha_2 e^{-t/\tau}$ with $\alpha_1=10^5$, $\alpha_2=1.6\times10^6$, and $\tau=1.5$. Left: optimal control $C(t)$ for anti-amyloid-$\beta$ and anti-tau treatment. 
Right: resulting percentage reduction. 
In both cases, treatment significantly suppresses amyloid-$\beta$ and tau progression compared to the untreated trajectories.
}
    \label{fig:population_treatment}
\end{figure*}

In the first experiment, we consider a constant penalty weight $\alpha$ in~\eqref{eq:cost} with $\alpha = 4\times 10^5$.
We then repeat the analysis using a time-dependent penalty function
\begin{equation}\label{eq:decaying-alpha-penalty}
    \alpha(t) = \alpha_1 + \alpha_2 e^{-t / \tau},
\end{equation}
with $\alpha_1 = 10^5$, $\alpha_2 = 1.6\times 10^6$, and $\tau = 1.5$.
The resulting optimal treatment strategies and corresponding system responses are shown in Fig.~\ref{fig:population_treatment}.

\subsection{Patient-wise Optimal Control}
\label{subsec:control-numeric-patientwise}

\begin{figure*}[h]
    \centering
    \includegraphics[width=0.6\linewidth]{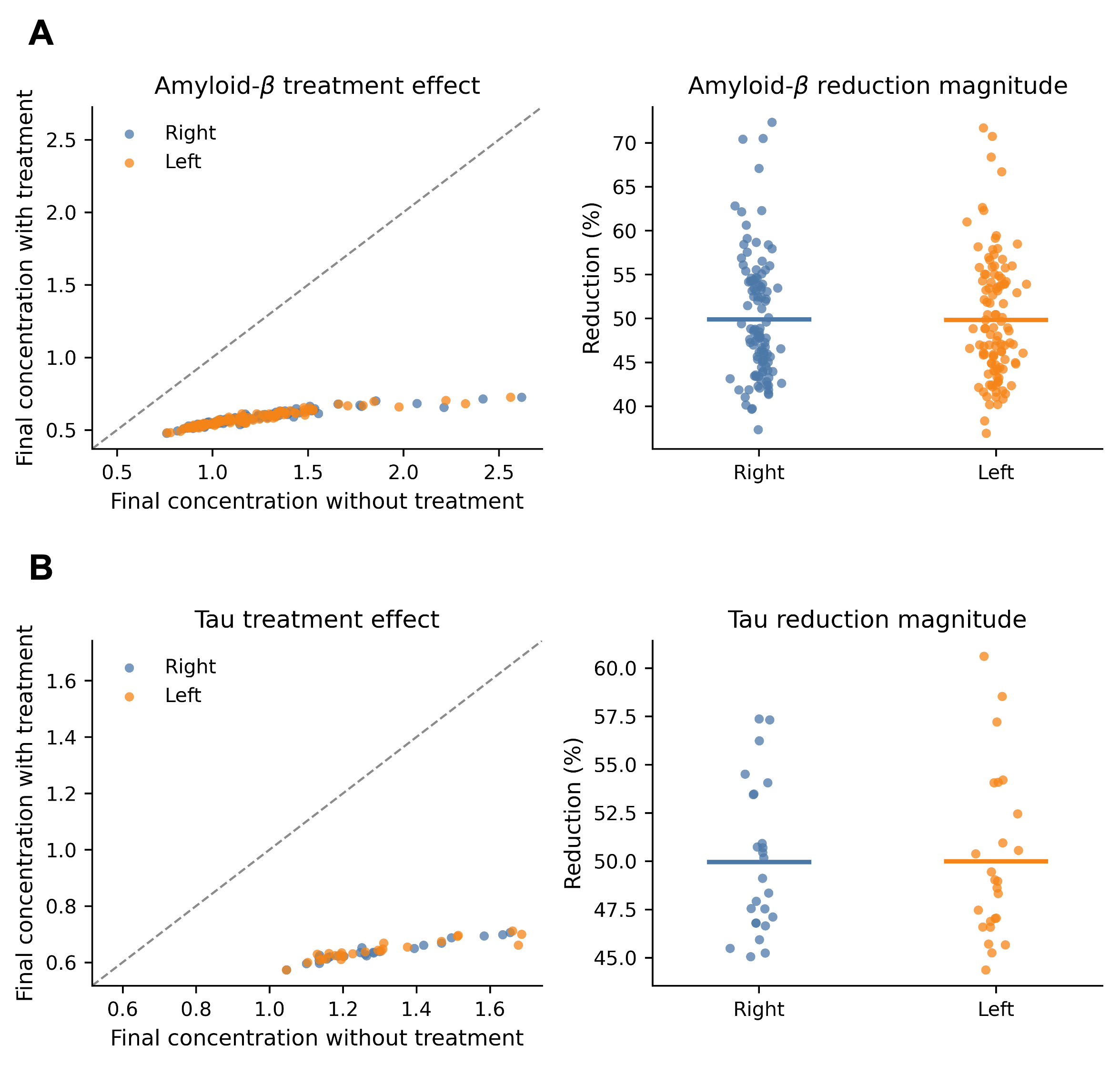}
    \caption{
\bf{Patient-specific treatment effects predicted by the learned digital twin for amyloid-$\beta$ (top, 111 subjects) and tau (bottom, 25 subjects).} 
Left: Final biomarker concentration with treatment versus without treatment for individual subjects in the left and right hemispheres. 
The dashed line indicates equality; points below the line represent treatment-induced reduction. 
Right: Distribution of percentage reductions across subjects, with horizontal bars indicating cohort means. 
Reductions are consistently observed across both hemispheres and biomarkers, with moderate inter-subject variability.
}
    \label{fig:patientwise_results}
\end{figure*}

In the patient-specific optimal control experiments, we extend the proposed framework to individualized settings for both amyloid-$\beta$ and tau by initializing the state equation with the observed biomarker distribution at the first available time point for each subject.
To model treatment-associated side effects, we adopt the time-dependent penalty function~\eqref{eq:decaying-alpha-penalty}.
For each patient, we compute two trajectories: one under the optimal treatment strategy and one without treatment ($C \equiv 0$).
The resulting trajectories are compared in terms of final biomarker burden to quantify the predicted treatment effect, as shown in Fig.~\ref{fig:patientwise_results}.
All patient-specific experiments are solved using the forward--backward sweep method described in \textcolor{blue}{SI Appendix 6}.

\subsection{Enabling immersive visualization and interaction}
We develop a virtual reality (VR) platform that provides an intuitive interface for interacting with the brain digital twin (DT). 
Using a six degree-of-freedom (6-DoF) interaction paradigm, users can manipulate the model through rotation, translation, and scaling to examine biomarker accumulation in specific brain regions. Such immersive visualization facilitates the exploration of disease-related spatial patterns on the complex three-dimensional cortical geometry.

The platform also supports longitudinal exploration through an interactive temporal control interface. By adjusting an age slider, users can navigate simulated disease trajectories and visualize the spatiotemporal evolution of biomarkers over time. In addition, treatment scenarios can be compared with untreated disease progression, allowing users to examine the projected effects of interventions on biomarker propagation. These capabilities provide a framework for visualizing model predictions and facilitating the interpretation of disease progression and treatment outcomes.

\begin{figure*}[htbp]
    \centering
    \includegraphics[width=0.85\linewidth]{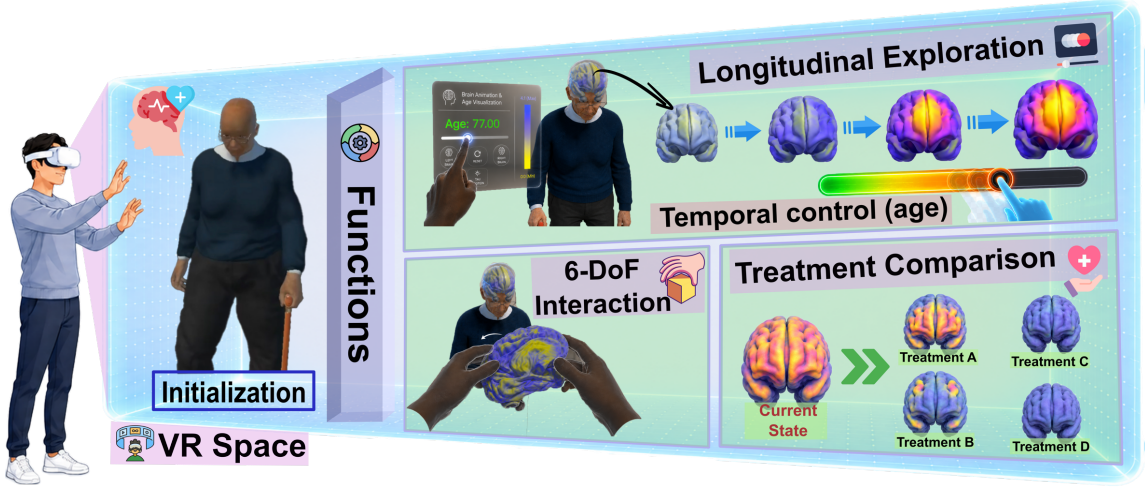}
\caption{ \textbf{Virtual reality platform for interactive visualization of the brain digital twin.} The platform enables immersive exploration of patient-specific biomarker distributions and disease trajectories on cortical surfaces. It supports six degree-of-freedom (6-DoF) interaction, longitudinal navigation of disease evolution through a temporal interface, and comparison of predicted treatment and untreated outcomes within a unified visual environment. }
    \label{fig:vr-platform}
\end{figure*}

Technical details of the VR platform are provided in \textcolor{blue}{SI Appendix 8}, and its main functionalities are also illustrated in Figure~\ref{fig:vr-platform}.

\section{Discussion}

We develop a data-driven digital twin framework that integrates reaction--diffusion modeling on cortical surfaces, operator learning from longitudinal neuroimaging data, PDE-constrained optimal control for treatment design, and immersive visualization for interactive exploration of disease progression.

The proposed framework combines forward prediction of biomarker evolution with inverse optimization of therapeutic interventions. In the forward component, a Laplacian Eigenfunction-Based Neural Operator (LENO) is used to learn nonlinear reaction--diffusion dynamics directly from longitudinal PET imaging data. The resulting model demonstrates robust performance under sparse sampling and measurement noise, achieving predictive accuracies of 87\% for amyloid-$\beta$ and 81\% for tau. In the inverse component, treatment design is formulated as a PDE-constrained optimal control problem on the learned dynamics, enabling systematic adaptation of dosing strategies under different assumptions on treatment cost and side effects, and leading to substantial reductions in biomarker burden relative to untreated progression.

Beyond prediction and control, we provide an interactive virtual reality (VR) platform to support personalized digital twin modeling for individual Alzheimer's disease (AD) patients. The platform enables immersive exploration of patient-specific biomarker dynamics on cortical surfaces, including six degree-of-freedom manipulation, longitudinal navigation of disease trajectories, and comparison of simulated treatment and untreated outcomes. This environment facilitates direct engagement with individualized model predictions and supports qualitative assessment of disease progression in a patient-specific setting.

At a conceptual level, the framework provides a unified representation of disease progression and intervention that may support future clinical interpretation of Alzheimer’s disease dynamics. At the same time, several important limitations and extensions remain.

A key biological consideration is that amyloid-$\beta$ and tau are coupled pathological processes in Alzheimer’s disease. In this study, we model them independently due to the lack of sufficiently paired longitudinal measurements across both modalities within the available datasets. As a result, the coupled reaction dynamics are not identifiable from the current data. This limitation is therefore data-driven rather than methodological: the proposed operator learning framework naturally extends to vector-valued and coupled systems when appropriate multimodal longitudinal datasets become available.

Similarly, in the optimal control formulation, treatment side effects are represented through prescribed penalty functions to demonstrate feasibility of personalized intervention design. In realistic clinical settings, such penalty terms should themselves be inferred from data, reflecting patient-specific risk profiles and treatment tolerance. Incorporating data-driven identification of these cost functionals represents an important direction for future work.

Overall, this study demonstrates the potential of combining operator learning, mechanistic modeling, optimal control, and immersive visualization to construct interpretable digital twins for neurodegenerative disease, and suggests a path toward data-informed, patient-specific prediction and intervention in complex biological systems.

\section{Materials and Methods}
\label{sec:materials-methods}

\subsection{Data description }
We employ PET scans from ADNI dataset \cite{mueller2005alzheimer} (\url{https://adni.loni.usc.edu}). 
The data are available to qualified investigators through the LONI \href{https://ida.loni.usc.edu/login.jsp}{Image and Data Archive (IDA)}. 
Tau and Amyloid PET images were preprocessed using a structural MRI--guided pipeline, including PET--MRI registration, SUVR normalization using the cerebellar cortex as reference, and projection onto the cortical surface. 
The resulting surface maps were resampled to the FreeSurfer \textit{fsaverage} template for cross-subject analysis. 
A complete description of the preprocessing pipeline, parameter values, and scripts to reproduce the results are provided in our GitHub repository (\url{https://github.com/2lineok/Optimal-Control/tree/main/PET_surface_pipeline}).
Additional details on the longitudinal PET cohorts, including scan-count distributions, spectral projection-error analysis, and subject exclusion criteria based on projection accuracy, are provided in \textcolor{blue}{SI Appendix 1}.

\subsection{Learning the nonlinear dynamics: LENO}
To approximate the nonlinear reaction operator in \eqref{eq:model_strong_form}, we adopt the Laplacian Eigenfunction-Based Neural Operator (LENO) \cite{wang2025laplacian}. 
The key is to project both the data and the governing PDE onto a truncated eigenspace of the Laplace–Beltrami operator defined on the cortical surface, and then parametrize the nonlinear reaction term through a neural network. 
The resulting reduced model is then fitted to the data using gradient-based optimization to learn the nonlinear reaction operator.

The corresponding weak form of the reaction--diffusion equation \eqref{eq:model_strong_form} is: find $u \in L^2(0,T;H^1(\Gamma))$ such that
\begin{equation}\label{eq:model_weak_form}
(u_t, v)_{L^2(\Gamma)} + (\nabla_\Gamma u, \nabla_\Gamma v)_{L^2(\Gamma)} 
= (\mathcal{F}(u), v)_{L^2(\Gamma)}
\end{equation}
for all $v\in H^1(\Gamma)$. 

We begin by computing the eigenvalues and eigenfunctions of the Laplace–Beltrami operator:
\begin{equation}
    -\Delta_\Gamma \phi = \lambda \phi. 
\end{equation}
The problem is discretized using the finite element method with piecewise linear functions on the cortical surface mesh.

 After discretization, we assemble the generalized eigenvalue problem
\begin{equation} \label{eq:eigenvalue-problem}
    A u = \lambda M u,
\end{equation}
where $A, M \in \mathbb{R}^{J \times J}$ are the stiffness and mass matrices, respectively, and $u \in \mathbb{R}^{J}$ is the eigenvector. 
The dimension $J$ corresponds to the number of degrees of freedom of the cortical surface mesh.
 We solve \eqref{eq:eigenvalue-problem} and and retain the first $P$ eigenpairs $(\lambda_i, \phi_i)$ ordered by increasing eigenvalue to form a truncated eigenspace: 
 \begin{equation}
 	V_P = \text{span}\{ \phi_i \}_{i=1}^P.
 \end{equation}
  The eigenvectors $\phi_i \in \mathbb{R}^J$  are orthonormal under the discrete $L^2$ inner product induced by the mass matrix  $(\cdot,\cdot)_M$.

The Laplacian Eigenfunction-Based Neural Operator parameterized by $\theta$ takes the form
\begin{equation}\label{eq:LENO}
    \mathcal{N}(\beta(u);\theta) = \sum_{i = 1}^P \mathcal{G}_i(\beta(u); \theta)\phi_i(x),
\end{equation}
where each $\mathcal{G}_i(u;\theta)$ is the $i$-th output component of a fully connected neural network $\mathcal{G}:\mathbb{R}^P \to \mathbb{R}^P$. 
The basis functions $\phi_i$ are the eigenfunctions of the Laplace-Beltrami operator. 

We now introduce the learning problem for the data-driven discovery of the nonlinear reaction operator. 
We will project both the data and the equation to the truncated eigenspace
\begin{equation}
    V_P = \text{span}\{ \phi_i \}_{i=1}^P. 
\end{equation}

Given observed data samples $\{u^n(x) , t_n\}_{n=1}^N$, we define the projection coefficients in $V_P$ by
\[
    \beta_i^n(u) = (u^n(x), \phi_i(x))_M, \quad i=1,\dots,P, 
\]
where $(\cdot,\cdot)_M$ denotes the discrete $L^2$ inner product induced by the mass matrix.

The numerical solution at time $t_n$ is approximated as
\[
    \tilde{u}^n(x) = \sum_{i=1}^P \tilde{\beta}_i^n \phi_i(x).
\]
Although fully implicit schemes offer superior stability for stiff nonlinear PDEs \cite{xu2023lack, hao2025stability}, they require solving a nonlinear neural-operator system at each time step. 
To reduce the computational cost during training, we instead employ a semi-implicit discretization \cite{shen2010numerical}, treating the diffusion term implicitly and the nonlinear term explicitly.
Restricting the variational  problem \eqref{eq:model_weak_form} to the eigenspace yields
\begin{equation}\label{eq:numerical-scheme-beta}
    \frac{\tilde{\beta}^n - \tilde{\beta}^{n-1}}{t_n - t_{n-1}} + \Lambda_P \tilde{\beta}^n = \tilde{f},
\end{equation}
where $\Lambda_P = \mathrm{diag}(\lambda_1,\dots,\lambda_P)$ and $\tilde{f}_i = (\mathcal{F}(\tilde{u}^{n-1}), \phi_i)$. 
We parametrize the unknown reaction term $\mathcal{F}$ by the neural operator \eqref{eq:LENO}. 
This leads to
\begin{equation}\label{eq:evolution-beta}
    \frac{\tilde{\beta}^n - \tilde{\beta}^{n-1}}{t_n - t_{n-1}} + \Lambda_P \tilde{\beta}^n 
    = \mathcal{G}( \tilde{\beta}^{n-1};\theta), 
\end{equation}
where $\mathcal{G}: \mathbb{R}^P \to \mathbb{R}^P $ is a neural network function parameterized by $\theta$.

We now formulate the loss function to learn the nonlinear reaction operator by jointly minimizing the data mismatch and the residual of the discretized PDE in the truncated eigenspace. 

To this end, we define the following quantity: 
\begin{equation}
    \mathcal{R}^n := \frac{\beta^n- \beta^{n-1}}{t_n - t_{n-1}} + D \Lambda_P \beta^n,
\end{equation}
where $\beta^{n-1}$ and $\beta^n$ are projection coefficients of observed data samples. 

The total loss is composed of two parts:
\begin{itemize}
    \item \emph{Data loss}, which enforces consistency between observed data and predicted data:
    \begin{equation}
        L^D(\theta) = \frac{1}{N} \sum_{n=1}^N \frac{\|\tilde{\beta}^n - \beta^n\|_{\ell^2}}{\|\beta^n\|_{\ell^2}}.
    \end{equation}
    Here, $\beta^n$ is the given data and $\tilde{\beta}^n$ is the model-predicted discrete solution. 
    \item \emph{Residual loss}, which ensures consistency with the discretized PDE:
    \begin{equation}
        L^R(\theta) = \frac{1}{N} \sum_{n=1}^N \frac{\|\mathcal{R}^n - \mathcal{G}(\beta^{n-1};\theta)\|_{\ell^2}}{\|\mathcal{R}^n\|_{\ell^2}}.
    \end{equation}
    It is defined by computing the left-hand side of the discretized PDE using the observed data, and matching it with the output of the neural network.
\end{itemize}
Finally, the neural operator is trained by minimizing the combined loss
\begin{equation}
    L(\theta) = L^D(\theta) + L^R(\theta). 
\end{equation}

\begin{figure*}
    \centering
    \includegraphics[width=0.9\linewidth]{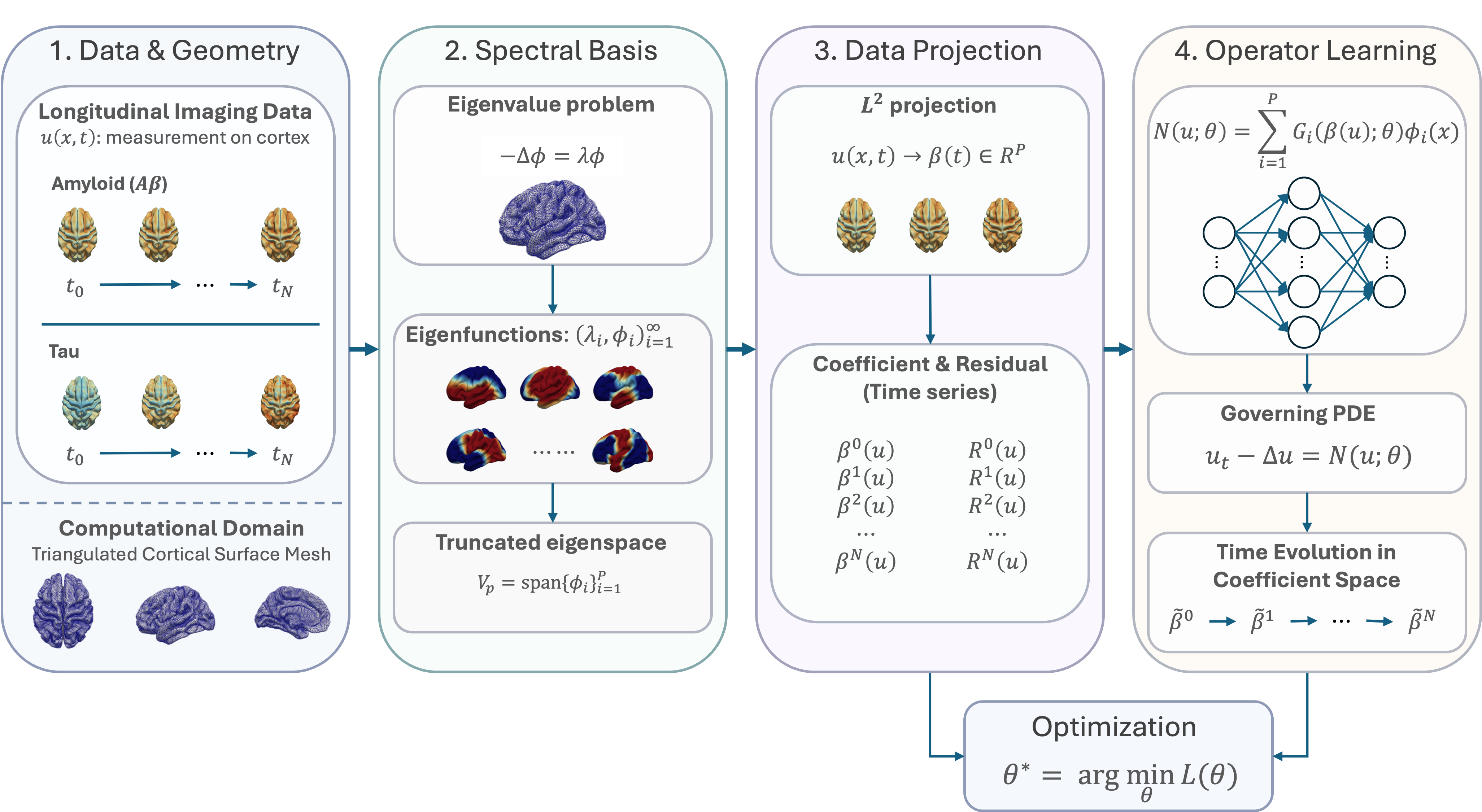}
\caption{
\textbf{Neural operator learning framework for cortical biomarker dynamics.}
Longitudinal amyloid-$\beta$ and tau measurements on the cortical surface are represented in a truncated Laplace--Beltrami eigenspace, yielding spectral coefficient trajectories. A neural operator is then trained to learn the nonlinear reaction operator governing biomarker evolution within a reaction--diffusion model from these data-driven representations. The resulting model enables inference of patient-specific disease dynamics from sparse longitudinal observations.
}    \label{fig:leno-workflow}
\end{figure*}

The entire workflow is illustrated in Figure \ref{fig:leno-workflow}. 

    In the training, we introduce a trainable time scaling parameter $\alpha$ in \eqref{eq:evolution-beta} for each patient, initially set to $1$. 
    
For model evaluation on prediction accuracy, we use patients with at least three time points. 
For these patients, all but the final time point are used for training, and the last available scan is reserved for evaluation of prediction accuracy.
 Other patients with only two time points are used exclusively for training. 
The relative $L^2$ errors in the eigenfunction coefficients and the nodal values are our primary evaluation metrics throughout; they are denoted by $E_1$ and $E_2$, respectively.
\begin{equation}
    E_1 = \frac{\|\beta - \hat{\beta}\|_{\ell_2}}{\|\beta\|_{\ell_2}}, \quad  E_2 = \frac{\|u - \hat{u}\|_{\ell_2}}{\|u\|_{\ell_2}},
\end{equation}
where $\beta$ and $\hat{\beta}$ denote the true and predicted coefficient vectors, respectively, and $u$ and $\hat{u}$ denote the true and predicted nodal values.

Prediction accuracy in the reduced eigenspace and on the cortical surface is then defined as
\begin{equation}
    \mathrm{Acc}_1
    =
    (1 - E_1) \times 100\%, \quad   \mathrm{Acc}_2
    =
    (1 - E_2) \times 100\%. 
\end{equation}

\subsection{Optimal control for treatment design}\label{sec:appendix-optimal-control}

We now use the learned disease dynamics to explore treatment strategies within a reduced-order control framework. 
We consider the reduced-order dynamics in the Laplacian eigenspace
\[
u(x,t)\approx \sum_{i=1}^P \beta_i(t)\phi_i(x),
\]
where $\{-\Delta\phi_i=\lambda_i\phi_i\}_{i=1}^P$ are the Laplacian eigenfunctions.
The coefficient vector $\beta(t)\in\mathbb R^P$ satisfies
\begin{equation}\label{eq:state_cont}
\dot\beta(t)
= -D\Lambda\beta(t) + \mathcal{G}_\theta(\beta(t)) - C(t)\beta(t),
\quad \beta(0)=\beta_0,
\end{equation}
where $\Lambda=\mathrm{diag}(\lambda_1,\dots,\lambda_P)$ and
$\mathcal{G}_\theta:\mathbb R^P\to\mathbb R^P$ is a learned nonlinear reaction operator
parameterized by a neural network as defined in \eqref{eq:LENO}.
Here $C(t)$ is the dosing intensity of the treatment at time $t$.

We seek an optimal dosing function $C(t)$ that minimizes
\begin{equation}
J(\beta,C)
=
\int_0^T \big(\|\beta(t)\|_2^2 + \alpha(t) C(t)^2\big)\,dt,
\end{equation}
subject to the dynamics \eqref{eq:state_cont}.
The resulting adjoint system is given by 
\begin{equation}\label{eq:adjoint_cont_beta}
\begin{cases}
-\dot{p}(t)
=
\big(
-D\Lambda
+
J_\mathcal{G}(\beta(t))
-
C(t)I
\big)^\top p(t)
- 2\beta(t),\\[2mm]
p(T)=0, \qquad t\in(0,T). 
\end{cases}
\end{equation}
Here, $p(t)$ is the Lagrange multiplier. $J_\mathcal{G}(\beta)$ denotes the Jacobian of $\mathcal{G}_\theta$. 
The first-order optimality condition is
\begin{equation}\label{eq:optimality_cont_beta}
2\alpha(t) C^*(t) + p(t)^\top \beta(t) = 0,
\quad
C^*(t)
=
-\frac{p(t)^\top \beta(t)}{2\alpha(t)}.
\end{equation}
We then adopt a forward--backward sweep method to solve the optimal control problem. 
The full derivation of the above optimality system and the algorithm are provided in \textcolor{blue}{SI Appendix 6}. 





\section{Data Availability}
The source code and processed data necessary to reproduce the findings of this study will be made publicly available at \url{https://github.com/georgexxu/neural-operator-based-brain-digital-twins/} upon publication.

\acknow{XX and WH were supported by National Institute of General Medical Sciences through grant 1R35GM146894.  WH was also supported by NSF DMS-2533995 and the Huck Chair in AI Mathematical Modeling from Penn State University's Huck Institutes of the Life Sciences.
}

\appendix

\section*{SI Appendix}

\setcounter{section}{0}

\renewcommand{\thesection}{\arabic{section}} 
\section{Data preprocessing and spectral representation}

\begin{figure}[h]
	\centering
	\begin{subfigure}[t]{0.49\linewidth}
		\centering
		\includegraphics[width=\linewidth]{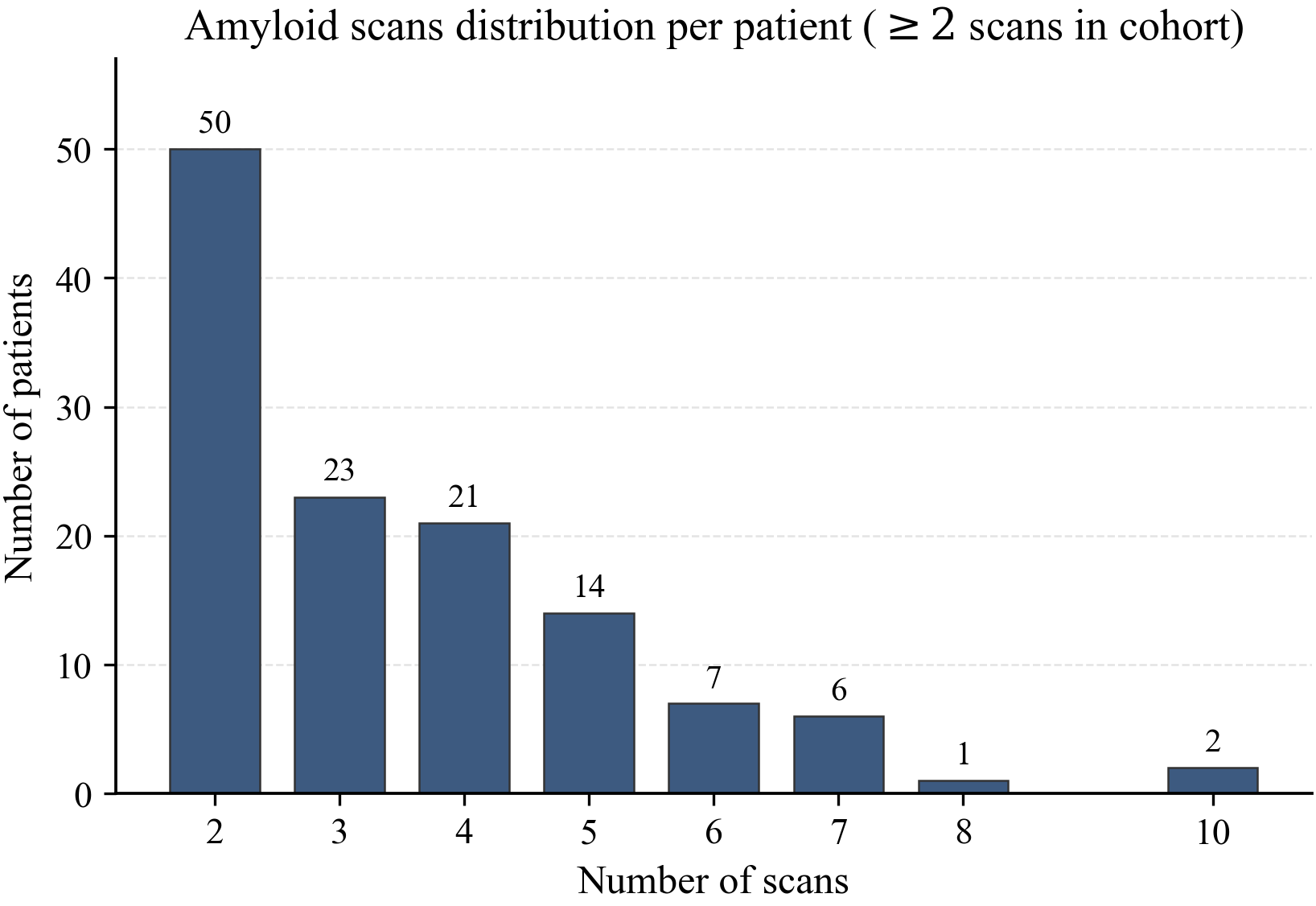}
		\label{fig:abeta-scan-count-left}
	\end{subfigure}\hfill
	\begin{subfigure}[t]{0.49\linewidth}
		\centering
		\includegraphics[width=\linewidth]{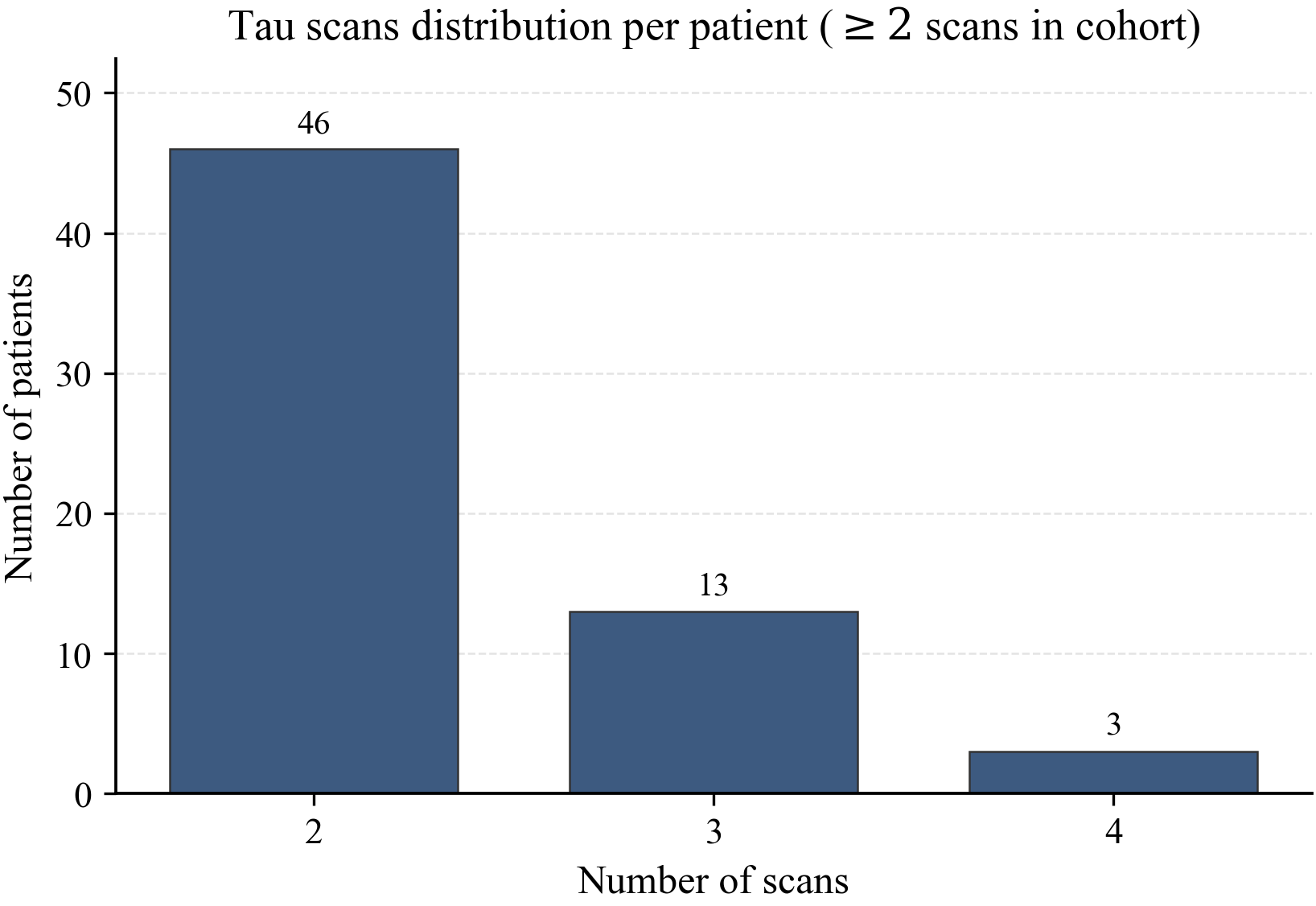}
		\label{fig:tau-scan-count-left}
	\end{subfigure}
	\caption{Longitudinal sampling of surface PET data after preprocessing. Amyloid-$\beta$ yields a larger, more densely sampled cohort than tau under the pipelines used here.}
	\label{fig:amyloid-tau-scan-count-left}
\end{figure}

Figure~\ref{fig:amyloid-tau-scan-count-left} shows the distribution of longitudinal scans per subject for amyloid-$\beta$ and tau after preprocessing.
The amyloid-$\beta$ cohort is larger ($124$ subjects) and more densely sampled, with many subjects having three or more scans. 
In contrast, the tau cohort ($62$ subjects) is dominated by subjects with only two scans, indicating more limited temporal sampling. 
In both datasets, the irregular and sparse longitudinal sampling, together with inter-subject variability and noise in the measurements, introduces challenges for learning robust and generalizable dynamics.

\begin{figure}[h] 
	\centering 
	\begin{subfigure}[t]{0.46\linewidth} 
		\centering 
		\includegraphics[width=\linewidth]{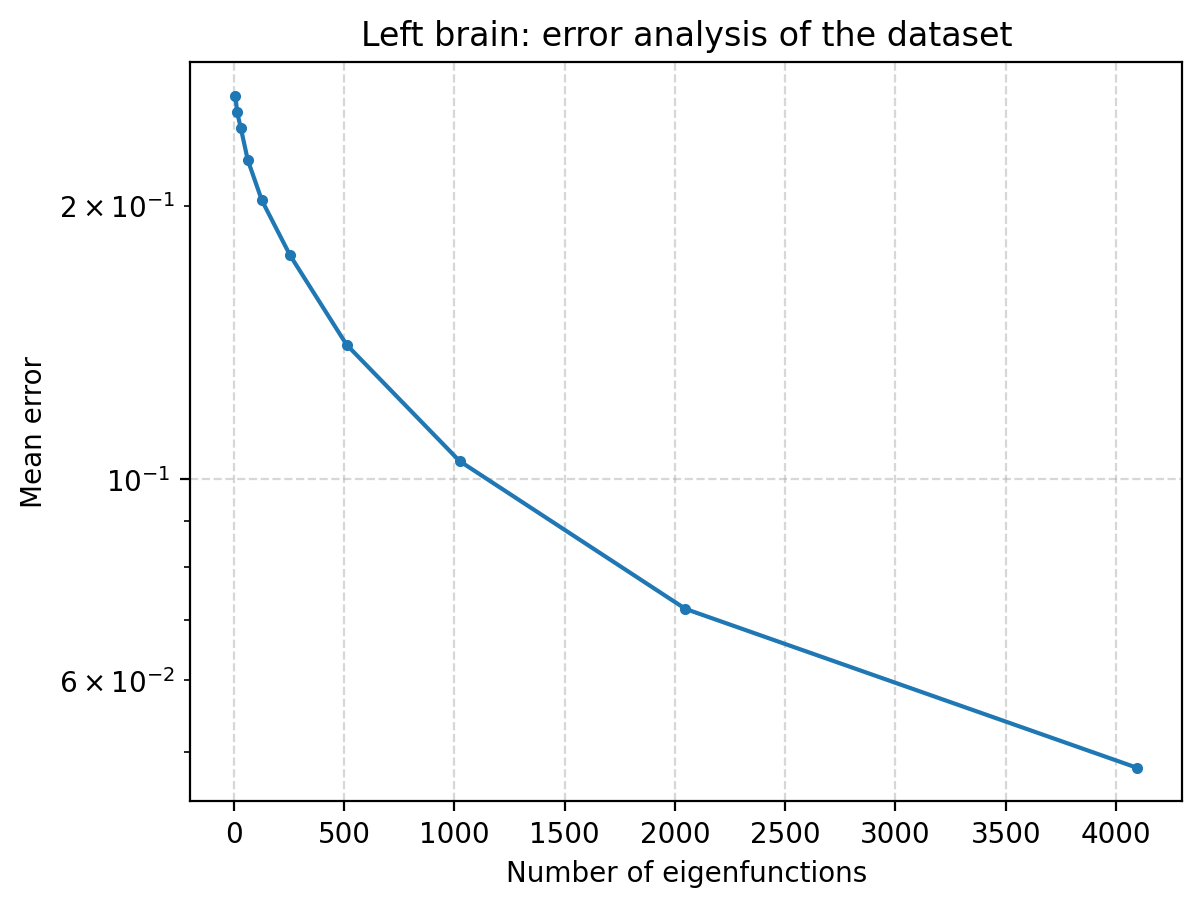} \caption{Relative $L^2$ projection error versus $p$, averaged over the amyloid-$\beta$ surface scans in the cohort.} \label{fig:brainl-err-dataset} 
	\end{subfigure}
	\hfill 
	\begin{subfigure}[t]{0.49\linewidth} \centering \includegraphics[width=\linewidth]{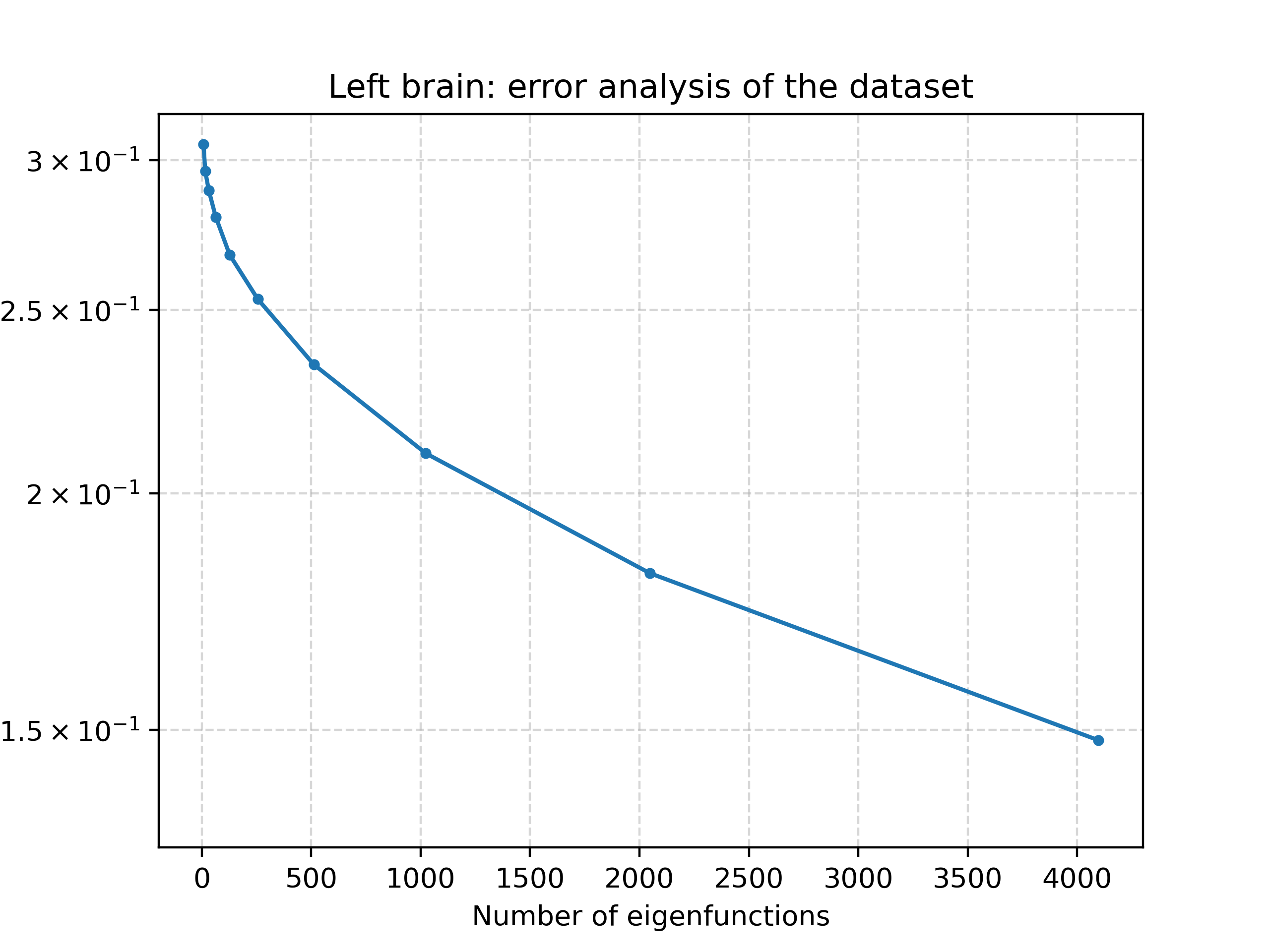} 
		\caption{Relative $L^2$ projection error versus $p$, averaged over the tau surface scans in the cohort.} \label{fig:tau-left-hemisphere-error}
	\end{subfigure} 
	\caption{Spectral projection-error analysis for left-hemisphere amyloid and tau data.} \label{fig:brainl-err-projection} 
\end{figure}

To represent the cortical fields, we project the data onto a truncated Laplace--Beltrami eigenspace. 
The choice of the truncation level $P$ is not fixed a priori, but is guided by a projection error analysis. 
We investigate how the truncation level $P$ of the Laplacian eigenspace affects the approximation quality of the data and then fileter out some data points that have large projection errors.
Figure~\ref{fig:brainl-err-projection} shows the relative $L^2$ projection error as a function of $P$ when projecting cortical amyloid-$\beta$ (Panel~(a)) and tau (Panel~(b)) fields onto the first $P$ Laplace--Beltrami eigenmodes. 
The results indicate a clear decay of the projection error as $P$ increases, suggesting that the dominant spatial patterns are well captured. 
With a total of 4096 eigenfunctions, the mean projection error over the entire dataset for the left-hemisphere amyloid-$\beta$ is less that 0.06 and for the left-hemisphere tau is less than 0.15. 
The larger projection error of tau data indicates that tau data is spatially more heterogeneous than amyloid-$\beta$ data.
Similar results are obtained for the right-hemisphere amyloid-$\beta$ and tau data. 

Based on this error analysis, we restrict our experiments to $P \in \{1500, 2048, 4096\}$, which provides a balance between approximation accuracy and computational efficiency. 
Furthermore, we filter out subjects with large projection errors. In particular, we remove subjects whose projection error exceeds 0.1 for amyloid-$\beta$ and 0.15 for tau. 
This yields a total of 111 subjects for amyloid-$\beta$ and 25 subjects for tau data. 

\section{Supplementary experiments for amyloid-$\beta$ and tau data}
\label{sec:appendix-depth-search}

To assess the effect of spectral truncation, we test multiple choices of the number of Laplacian eigenfunctions, characterized by the truncation level $P \in \{1500, 2048, 4096\}$.
The prediction accuracies for the left and right hemispheres are summarized in Table~\ref{tab:exp-tanh-wd-1e-2-more-data} and Table~\ref{tab:right-exp-tanh-wd-1e-2-p2048-more-data}, respectively.

\begin{table}[H]
	\centering
	\begin{tabular}{|c|c|c|c|c|c|c|}\hline
		NN architecture&   \multicolumn{2}{|c|}{ $P=1500$}&\multicolumn{2}{|c|}{Tanh, $P=2048$} & \multicolumn{2}{|c|}{ $P=4096$}\\\hline
		$[P,M,P]$&  $\mathrm{Acc}_1$ &  $\mathrm{Acc}_2$ & $\mathrm{Acc}_1$ &  $\mathrm{Acc}_2$ & $\mathrm{Acc}_1$ &  $\mathrm{Acc}_2$ \\\hline
		$M = 256$  & 88.1\% & 86.1\% & 87.7\% & 86.5\% & 87.0\% & 86.6 \% \\\hline
		$M = 512$ & 88.2\% & 86.2\% & 87.9\% & 86.6\% & 87.5\% & 87.0 \% \\\hline
		$M = 1024$ & 88.3\% & 86.3\% & 88.0\% & 86.7\% & 87.7\% & 87.2 \% \\\hline
		$M =2048$ &  88.3\% & 86.3\% & 87.9\% & 86.6\% & 87.3\% & 87.2 \% \\ \hline
	\end{tabular}
	\caption{Prediction accuracy for amyloid-$\beta$ in the left hemisphere using one-hidden-layer $\tanh$ networks with weight decay $10^{-2}$. 
		Results are reported for different spectral truncation levels $P$ and network widths $M$.}
	\label{tab:exp-tanh-wd-1e-2-more-data}
\end{table}

\begin{table}[H]
	\centering
	\begin{tabular}{|c|c|c|c|c|c|c|}\hline
		NN architecture&\multicolumn{2}{|c|}{Tanh, $P=1500$} & \multicolumn{2}{|c|}{Tanh, $P=2048$} & \multicolumn{2}{|c|}{Tanh, $P=4096$}\\\hline
		$[P,M,P]$&$\mathrm{Acc}_1$&  $\mathrm{Acc}_2$ & $\mathrm{Acc}_1$&$\mathrm{Acc}_2$ & $\mathrm{Acc}_1$ & $\mathrm{Acc}_2$  \\\hline
		M = 256  & 88.5\%&  86.5\% & 88.1\% & 86.8\% & 87.3\% & 86.9\% \\\hline
		M = 512 & 88.6\% & 86.7\% & 88.3\%& 87.1\% & 87.9\% & 87.4\%  \\\hline
		M = 1024& 88.7\% & 86.7\% & 88.5\% & 87.2\%  &88.2\% & 87.7\% \\\hline
		M =2048 & 88.8\% & 86.8\% & 88.4\%& 87.1\% & 88.1\% &  87.6\%\\ \hline
	\end{tabular}
	\caption{Prediction accuracy for amyloid-$\beta$ in the right hemisphere using one-hidden-layer $\tanh$ networks with weight decay $10^{-2}$. 
		Results are reported for different spectral truncation levels $p$ and network widths $M$.}
	\label{tab:right-exp-tanh-wd-1e-2-p2048-more-data}
\end{table}

For both hemispheres, the best $\mathrm{Acc}_2$ is achieved at $p = 4096$ with $M = 1024$ or $M = 2048$. 
Our model achieves a prediction accuracy of 87.2\% in the left hemisphere and 87.6\% in the right hemisphere, measured by the $L^2$ error in the physical space.

Across all configurations, reducing the truncation level $P$ leads to comparable or slightly improved performance in $\mathrm{Acc}_1$, suggesting that learning the nonlinear dynamics in a lower-dimensional spectral space is more effective. 
However, $\mathrm{Acc}_2$, which measures the $L^2$ error in the physical space, does not consistently improve with smaller $P$, reflecting the loss of high-frequency information due to spectral truncation. 

In terms of model capacity, increasing the network width improves performance up to $M \approx 1024$--$2048$, beyond which the gains saturate, indicating that the approximation of the nonlinear reaction term is no longer the dominant source of error. 
Finally, both hemispheres exhibit highly consistent trends across all settings, supporting the robustness of these observations. 
Overall, the results highlight a trade-off: while reduced spectral representations facilitate optimization and improve learning efficiency, the ultimate accuracy in the physical domain is limited by the approximation error of the truncated eigenbasis.

We apply the LENO framework to learn the nonlinear reaction dynamics from tau data in the left hemisphere. To assess the effect of spectral truncation, we test multiple choices of the number of Laplacian eigenfunctions.

Within the LENO architecture, we employ a shallow fully connected neural network with \texttt{tanh} activation as the branch network to approximate the nonlinear reaction term in the reaction--diffusion equation.

\begin{table}[H]
	\centering
	\begin{tabular}{|c|c|c|c|c|c|c|}\hline
		NN architecture&\multicolumn{2}{|c|}{Tanh, $P=1500$} & \multicolumn{2}{|c|}{Tanh, $P=2048$} & \multicolumn{2}{|c|}{Tanh, $P=4096$}\\\hline
		$[P,M,P]$&$\mathrm{Acc}_1$&  $\mathrm{Acc}_2$ & $\mathrm{Acc}_1$&$\mathrm{Acc}_2$ & $\mathrm{Acc}_1$ & $\mathrm{Acc}_2$  \\\hline
		$M = 32$  &   84.2\%& 80.6\%  & 83.3\%& 80.5\% & 81.9\% & 80.5\%\\\hline
		$M = 64$  &   84.7 \%& 81.0\% & 84.1\%& 81.2\% &82.7\% & 81.3\%\\\hline
		$M = 128$ &  84.2 \%& 80.6\%  & 83.9\%& 81.0\% & 82.5\% & 81.1\% \\\hline
		$M = 256$ & 83.9\%  & 80.3\% & 83.2\%&  80.4\% & 82.1\% & 80.7\% \\\hline
	\end{tabular}
	\caption{Prediction accuracy for tau in the left hemisphere using one-hidden-layer $\tanh$ networks with weight decay $10^{-2}$. 
		Results are reported for different spectral truncation levels $p$ and network widths $M$.}
	\label{tab:left-tau-exp}
\end{table}

\begin{table}[H]
	\centering
	\begin{tabular}{|c|c|c|c|c|c|c|}\hline
		NN architecture&\multicolumn{2}{|c|}{Tanh, $P=1500$} & \multicolumn{2}{|c|}{Tanh, $P=2048$} & \multicolumn{2}{|c|}{Tanh, $P=4096$}\\\hline
		$[P,M,P]$&$\mathrm{Acc}_1$&  $\mathrm{Acc}_2$ & $\mathrm{Acc}_1$&$\mathrm{Acc}_2$ & $\mathrm{Acc}_1$ & $\mathrm{Acc}_2$  \\\hline
		M = 32  & 84.0\%  & 80.4\% & 83.1\% &  80.3\% & 81.7\% & 80.2\%\\\hline
		M = 64  & 84.5\% &80.8\%  & 83.8\% & 80.9\% & 82.4\% & 80.9\%\\\hline
		M = 128  &83.9\% & 80.3\% & 83.5\% & 80.8\% & 82.4\%& 81.1\%\\\hline
		M = 256 &83.6\% & 80.1\% &  81.5\% & 80.1\% &81.9\% & 80.5\%\\\hline
	\end{tabular}
	\caption{Prediction accuracy for tau in the right hemisphere using one-hidden-layer $\tanh$ networks with weight decay $10^{-2}$. 
		Results are reported for different spectral truncation levels $p$ and network widths $M$.}
	\label{tab:right-tau-exp}
\end{table}

For both hemispheres, the best $\mathrm{Acc}_2$ is achieved at $P = 4096$ with $M = 128$.  
Our model achieves a prediction accuracy of 81.3\% in the left hemisphere and 81.0\% in the right hemisphere, measured by the $L^2$ error in the physical space.

Reducing the truncation level $P$ leads to improved performance in $\mathrm{Acc}_1$, suggesting that learning the nonlinear dynamics in a lower-dimensional spectral space is more effective.
However, $\mathrm{Acc}_2$, which measures the $L^2$ error in the physical space, does not consistently improve with smaller $P$, reflecting the loss of high-frequency information due to spectral truncation.
In terms of model capacity, increasing the network width improves performance up to $M \approx 128$, beyond which the gains saturate, indicating that the approximation of the nonlinear reaction term is no longer the dominant source of error.
The results are consistent across hemispheres and with the amyloid-$\beta$ results.

For amyloid-$\beta$ data, we also conduct a grid search over network architectures and regularization parameters to assess the impact of model capacity on prediction accuracy, as summarized in Table~\ref{tab:hyperparameter-list}.
We train the neural networks for 500 epochs with an initial learning rate of 1e-3 that decays by half every 50 steps.
For each depth, we report the configurations that achieve the best performance in terms of $\mathrm{Acc}_1$ and $\mathrm{Acc}_2$. 
The results are presented in Tables~\ref{tab:depth2}--\ref{tab:depth5}.

Overall, tanh neural networks outperform ReLU neural networks in these experiments. 
Also, increasing network depth beyond two hidden layers does not lead to improvements in accuracy.
In fact, deeper architectures tend to exhibit slightly degraded performance, suggesting that the dominant sources of error arise from spectral truncation and data quality rather than model expressivity. 

\begin{table}[H]
	\centering
	\begin{tabular}{|c|c|}\hline
		\textbf{Hyperparameter} & \textbf{Values} \\  \hline
		Activation function&ReLU, tanh\\ \hline
		Weight decay& 1e-2, 5e-3, 1e-3, 5e-4\\\hline
		Network depth& 2, 3, 4, 5\\\hline
		Network width ($M$) & 128, 256, 512\\ \hline
		Spectral truncation ($P$) & 1500, 2048, 4096\\\hline
	\end{tabular}
	\caption{Hyperparameter explored in the grid search over activation functions, network architectures, regularization, and spectral truncation level.}
	\label{tab:hyperparameter-list}
\end{table}

\begin{table}[H]
	\centering
	\begin{tabular}{|c|c|c|c|c|c|c|}\hline
		NN architecture&  \multicolumn{2}{|c|}{Tanh, $P=1500$} & \multicolumn{2}{|c|}{Tanh, $P=2048$} & \multicolumn{2}{|c|}{Tanh, $P=4096$}\\\hline
		$[P,M^2,P]$&  $\mathrm{Acc}_1$ &  $\mathrm{Acc}_2$ & $\mathrm{Acc}_1$ &  $\mathrm{Acc}_2$ & $\mathrm{Acc}_1$ &  $\mathrm{Acc}_2$ \\\hline
		$M = 128$& 87.7\% & 85.8\% & 87.4\% & 86.2\% & 86.7\% &  86.3\% \\\hline
		$M = 256$& 87.9\% &86.0\% & 87.8\% & 86.5\% & 87.4\% & 86.9\% \\\hline
		$M = 512$& 87.9\% &86.0\%& 87.7\% & 86.4\% & 87.5\% & 87.0\%\\ \hline
	\end{tabular}
	\caption{Prediction accuracy ($\mathrm{Acc}_1$, $\mathrm{Acc}_2$) of fully connected neural networks with $2$ hidden layers ($[P, M^2, P]$) across different widths $M$ and spectral truncation levels $P$.}
	\label{tab:depth2}
\end{table}

\begin{table}[H]
	\centering
	\begin{tabular}{|c|c|c|c|c|c|c|}\hline
		NN architecture&  \multicolumn{2}{|c|}{Tanh, $P=1500$} & \multicolumn{2}{|c|}{Tanh, $P=2048$} & \multicolumn{2}{|c|}{Tanh, $P=4096$}\\\hline
		$[P,M^3,P]$&  $\mathrm{Acc}_1$ &  $\mathrm{Acc}_2$ & $\mathrm{Acc}_1$ &  $\mathrm{Acc}_2$ & $\mathrm{Acc}_1$ &  $\mathrm{Acc}_2$ \\\hline
		$M = 128$&  87.3\% & 85.5\% & 86.8\% & 85.6\% & 86.4\% & 86.0\% \\\hline
		$M = 256$& 87.6\% & 85.7\% & 87.4\% & 86.1\% & 87.0\% & 86.5\% \\\hline
		$M = 512$& 87.5\% & 85.6\% & 87.4\% & 86.2\% & 87.1\% & 86.7\% \\ \hline
	\end{tabular}
	\caption{Prediction accuracy ($\mathrm{Acc}_1$, $\mathrm{Acc}_2$) of fully connected neural networks with $3$ hidden layers ($[P, M^3, P]$) across different widths $M$ and spectral truncation levels $P$. }
	\label{tab:depth3}
\end{table}

\begin{table}[H]
	\centering
	\begin{tabular}{|c|c|c|c|c|c|c|}\hline
		NN architecture&  \multicolumn{2}{|c|}{Tanh, $P=1500$} & \multicolumn{2}{|c|}{Tanh, $P=2048$} & \multicolumn{2}{|c|}{Tanh, $P=4096$}\\\hline
		$[P,M^4,P]$&  $\mathrm{Acc}_1$ &  $\mathrm{Acc}_2$ & $\mathrm{Acc}_1$ &  $\mathrm{Acc}_2$ & $\mathrm{Acc}_1$ &  $\mathrm{Acc}_2$ \\\hline
		$M = 128$& 86.9\% & 85.1\% & 86.5\% & 85.3\%&  86.0\% & 85.6\%  \\\hline
		$M = 256$& 87.2\% & 85.4\% & 86.9\%  & 85.7\%  & 86.2\% & 85.8\% \\\hline
		$M = 512$& 87.3\% & 85.4\% &  87.2\% & 86.0\% & 86.9\% & 86.4\% \\ \hline
	\end{tabular}
	\caption{Prediction accuracy ($\mathrm{Acc}_1$, $\mathrm{Acc}_2$) of fully connected neural networks with $4$ hidden layers ($[P, M^4, P]$) across different widths $M$ and spectral truncation levels $P$.}
	\label{tab:depth4}
\end{table}

\begin{table}[H]
	\centering
	\begin{tabular}{|c|c|c|c|c|c|c|}\hline
		NN architecture&  \multicolumn{2}{|c|}{Tanh, $P=1500$} & \multicolumn{2}{|c|}{Tanh, $P=2048$} & \multicolumn{2}{|c|}{Tanh, $P=4096$}\\\hline
		$[P,M^5,P]$&  $\mathrm{Acc}_1$ &  $\mathrm{Acc}_2$ & $\mathrm{Acc}_1$ &  $\mathrm{Acc}_2$ & $\mathrm{Acc}_1$ &  $\mathrm{Acc}_2$ \\\hline
		$M = 128$&  86.6\% & 84.9\% &86.2  & 85.1\% & 85.7\% & 85.3\% \\\hline
		$M = 256$& 86.7\% & 84.9\% & 86.4\% & 85.3\% & 85.8\% &  85.4\% \\\hline
		$M = 512$& 86.7\% & 84.9\% & 86.5\% & 85.3\% & 86.0\% & 85.6\% \\ \hline
	\end{tabular}
	\caption{Prediction accuracy ($\mathrm{Acc}_1$, $\mathrm{Acc}_2$) of fully connected neural networks with $5$ hidden layers ($[P, M^5, P]$) across different widths $M$ and spectral truncation levels $P$.}
	\label{tab:depth5}
\end{table}

\section{Spectral analysis of simulated cortical biomarker fields}
\label{sec:appendix-spectral-analysis}
To quantify the spatial complexity of the simulated cortical biomarker distributions, we analyze their spectral decomposition in the Laplace--Beltrami eigenbasis of the cortical surface. Let $u(x,t)$ denote the biomarker field and $\{\phi_k\}$ the Laplace--Beltrami eigenmodes, which are orthonormal under the mass-matrix inner product. The spatial mean, corresponding to the constant mode $k=0$, is removed before analysis. The field is then expanded as
\begin{equation}
	u(x,t)=\sum_{k=1}^{P}\beta_k(t)\phi_k(x),
\end{equation}
where $\beta_k(t)=\langle u,\phi_k\rangle_M$ are the spectral coefficients. The spectral energy of mode $k$ is defined as
\begin{equation}
	E_k(t)=|\beta_k(t)|^2.
\end{equation}

We measure the concentration of spectral energy using the cumulative energy fraction
\begin{equation}
	C_K(t)=\frac{\sum_{k=1}^{K}E_k(t)}{\sum_{k=1}^{P}E_k(t)},
\end{equation}
which represents the proportion of total energy captured by the first $K$ nonconstant modes.
We also plot the energy density function for various grouped modes. 
We first divide the eigenmodes into various groups $\{ I_i\}$ and then the density is given by 
\begin{equation}
	\rho_i(t)=\frac{\sum_{k \in I_i } E_k(t)}{\sum_{k=1}^{P}E_k(t)}. 
\end{equation}

\begin{figure}[H]
	\centering
	\includegraphics[width=0.8\linewidth]{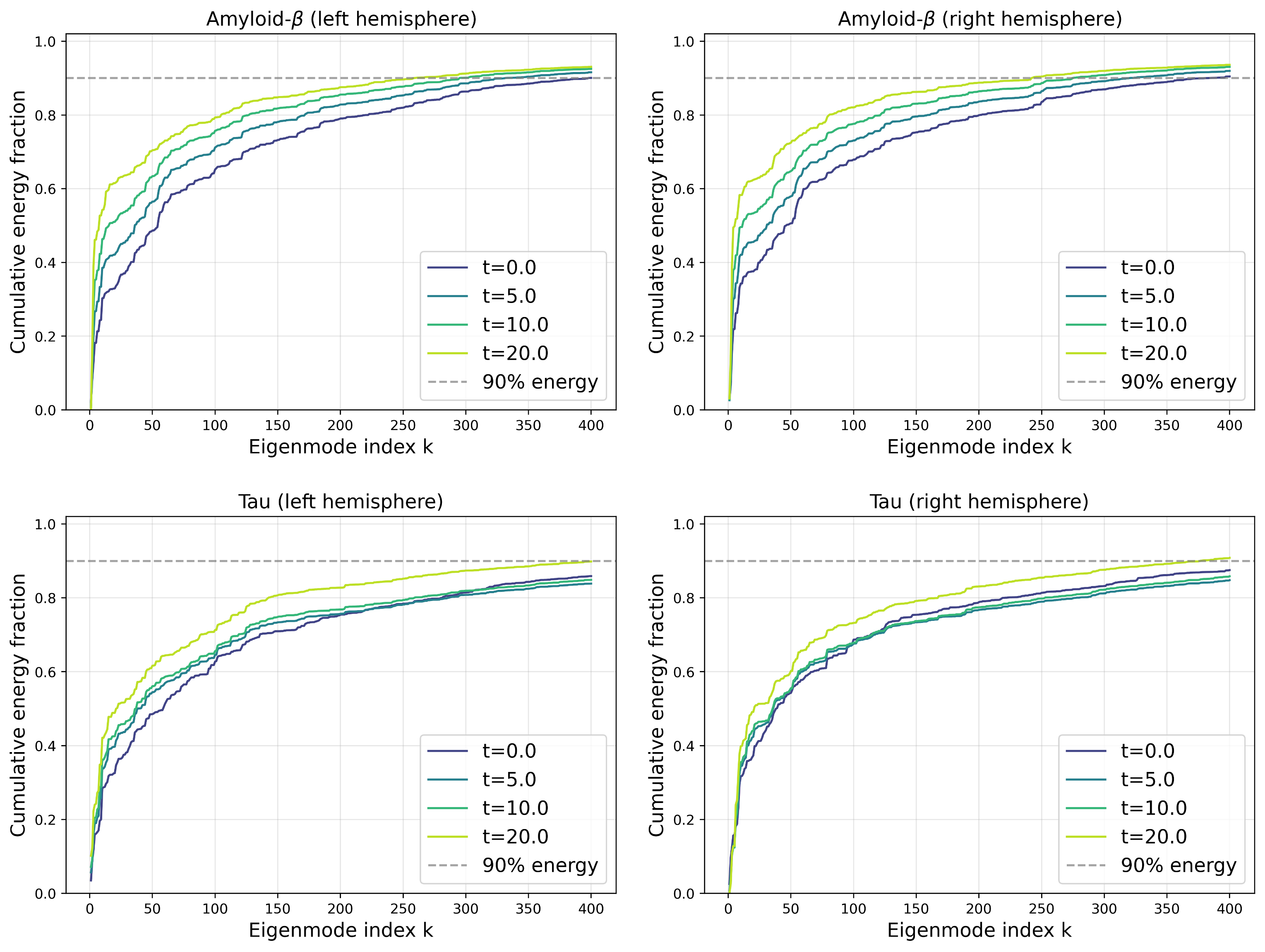}
	\caption{Cumulative spectral energy fractions of amyloid-$\beta$ (top) and tau (bottom) on the left (left column) and right (right column) hemispheres at $t=0,5,10,20$. }
	\label{fig:spectral_energy_cdf}
\end{figure}

\begin{figure}[H]
	\centering
	\includegraphics[width=0.8\linewidth]{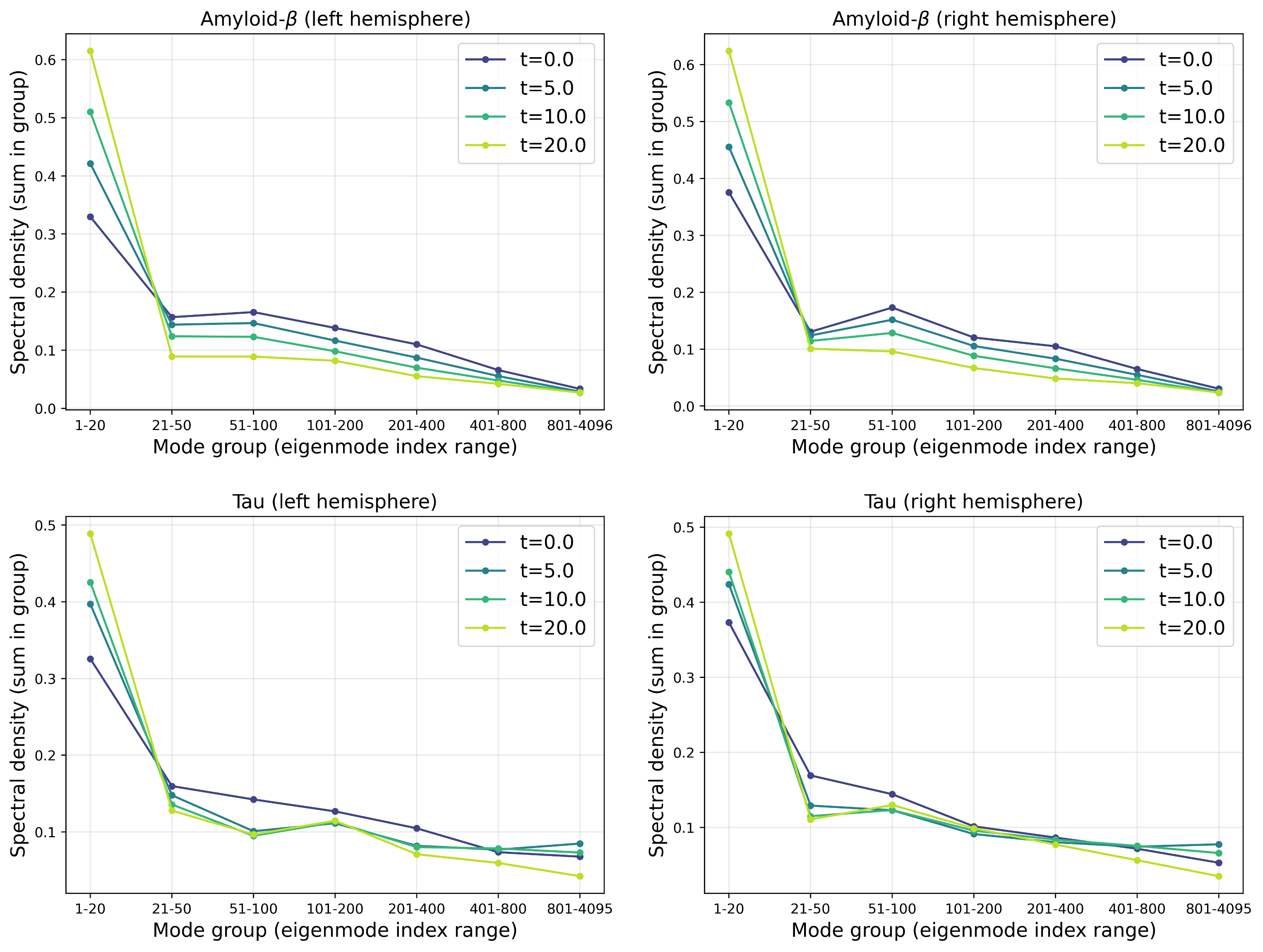}
	\caption{
		Grouped spectral density of amyloid-$\beta$ (top) and tau (bottom) on the left (left column) and right (right column) hemispheres at $t=0,5,10,20$. Each value represents the summed energy within a range of eigenmode indices.}
	\label{fig:spectral_energy_pdf}
\end{figure}

Figure~\ref{fig:spectral_energy_cdf} shows the cumulative spectral energy fractions for amyloid-$\beta$ and tau on both hemispheres at several time points. The amyloid-$\beta$ curves lie above the tau curves for comparable mode indices, indicating that a larger fraction of energy is captured by the lower Laplace--Beltrami modes. This suggests that the amyloid-$\beta$ distribution is spatially smoother and dominated by large-scale patterns. In contrast, the tau spectrum is more broadly distributed across modes, indicating richer spatial heterogeneity. In addition, the amyloid-$\beta$ curves shift upward over time, reflecting increasing concentration of energy in low-frequency modes, whereas tau exhibits a weaker and less monotonic temporal shift.

Figure~\ref{fig:spectral_energy_pdf} shows the grouped spectral energy distribution across different eigenmode bands. 
For amyloid-$\beta$, the majority of the energy is concentrated in the lowest-frequency band (modes 1--20), and this dominance becomes more pronounced over time, indicating increasingly smooth spatial patterns.
In contrast, tau exhibits a more distributed spectrum. The low-frequency band still dominates, but with relatively greater contributions from intermediate-frequency bands. reflecting greater spatial heterogeneity. 
This difference is consistent across both hemispheres. 
Moreover, while amyloid-$\beta$ shows a clear monotonic shift toward lower frequencies, tau displays weaker and less consistent temporal trends, suggesting more complex and less stable spectral dynamics.

\section{Evolution of the nonlinear reaction term}
\label{sec:appendix-evolution-nonlinear-reaction-term}
\begin{figure}[H]
	\centering
	\includegraphics[width=0.55\linewidth]{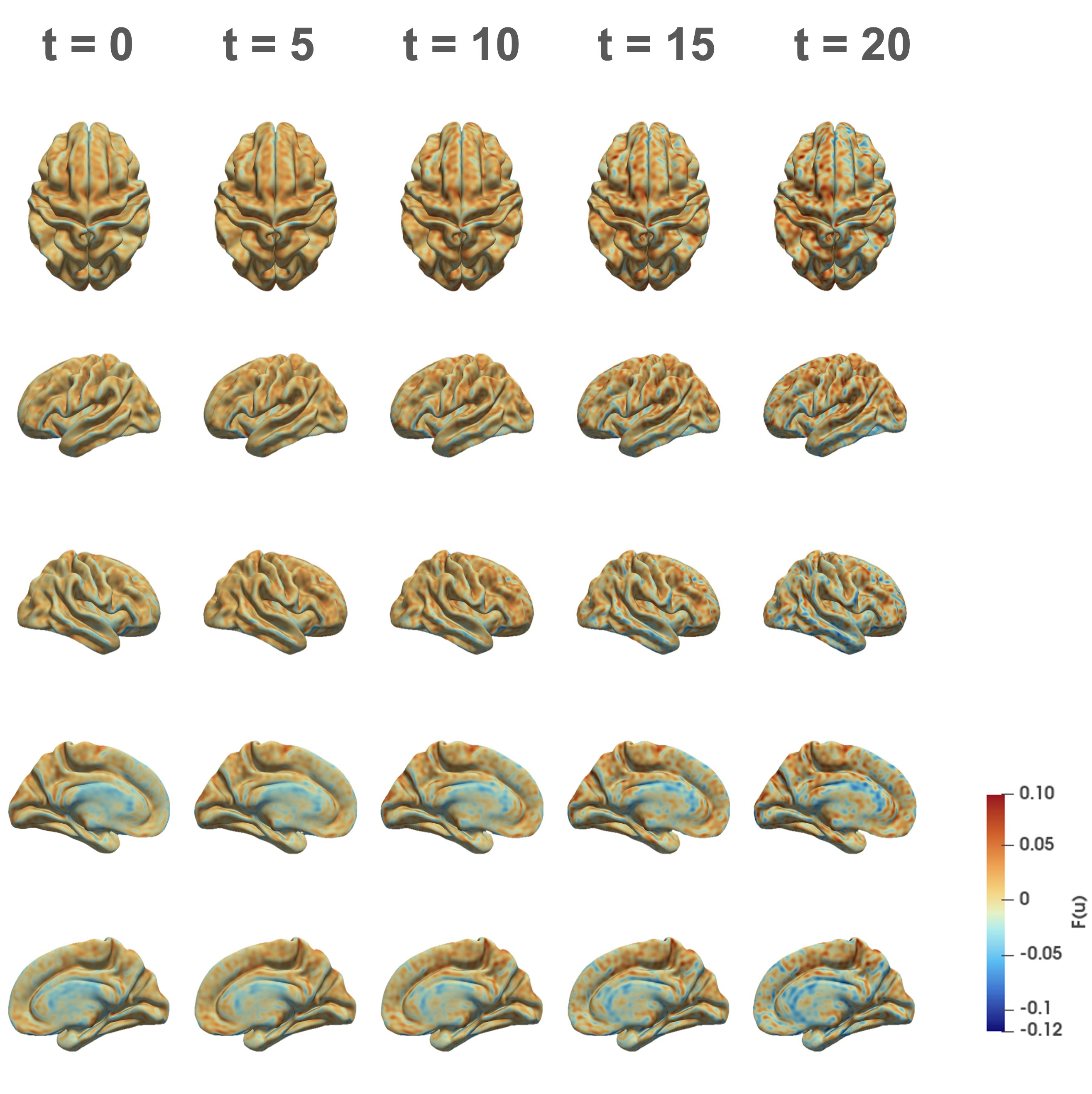}
	\caption{  Nonlinear reaction term in the population-level digital twin simulation of amyloid-$\beta$.
		The learned nonlinear term is evaluated along the simulated trajectory and visualized on the cortical surface.}
	\label{fig:abeta-nonlinear}
\end{figure}

\begin{figure}[H]
	\centering
	\includegraphics[width=0.55\linewidth]{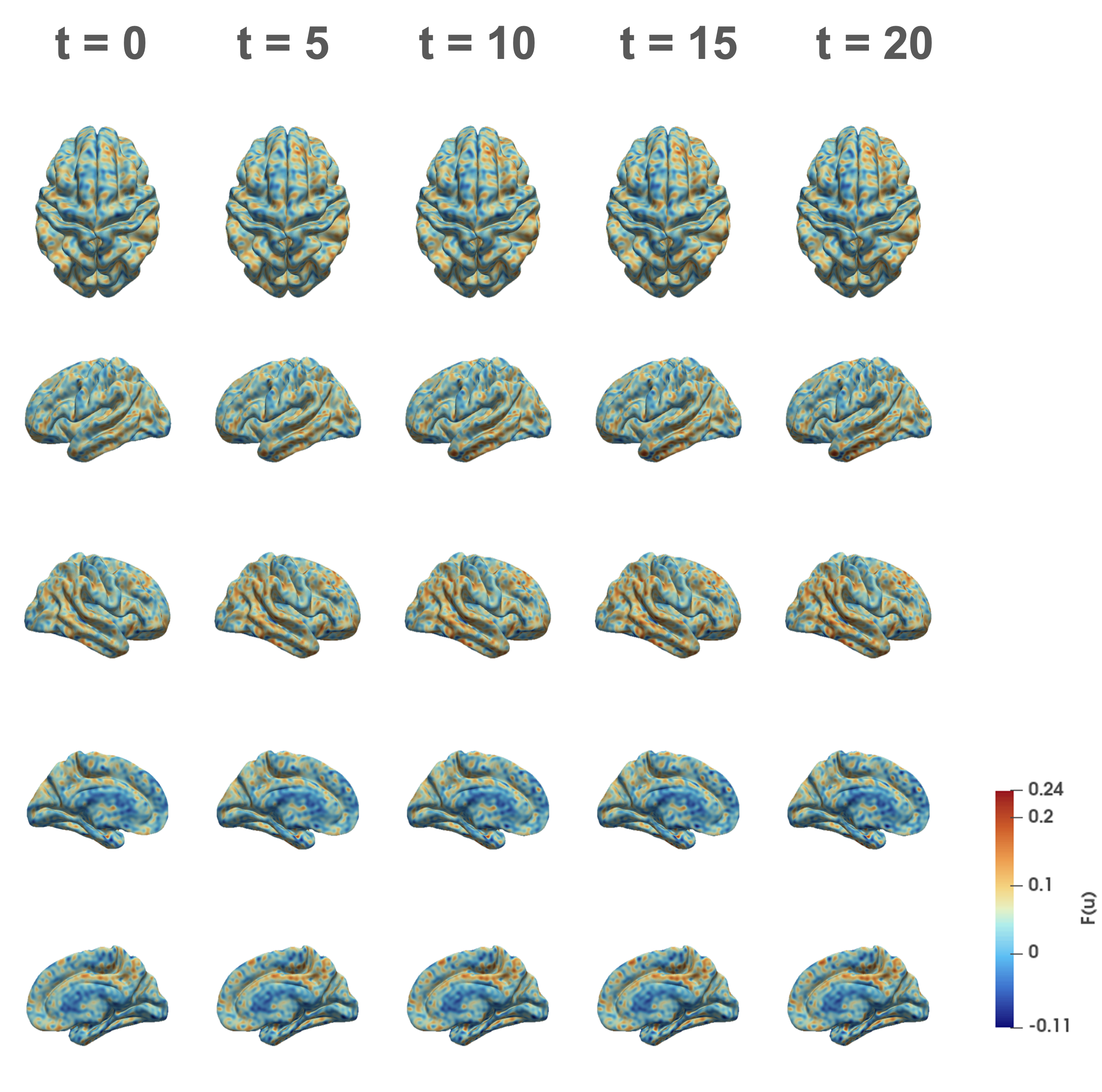}
	\caption{  Nonlinear reaction term in the population-level digital twin simulation of tau.
		The learned nonlinear term is evaluated along the simulated trajectory and visualized on the cortical surface.}
	\label{fig:tau-nonlinear}
\end{figure}

For the population-level digital twin simulation, we visualize the evolution of the nonlinear reaction term on the cortical surface for amyloid-$\beta$ and tau in Figures~\ref{fig:abeta-nonlinear} and~\ref{fig:tau-nonlinear}.

\section{Coupling Amyloid and Tau}
\label{sec:appendix-coupling-amyloid-tau}

We consider a coupled reaction--diffusion system describing the spatiotemporal
dynamics of amyloid-beta $u_A(x,t)$ and tau $u_T(x,t)$.
The amyloid-beta dynamics are assumed to be autonomous, whereas the evolution
of tau depends on both tau and amyloid-beta, inducing a unidirectional coupling
from amyloid-beta to tau.
The governing equations are given by
\begin{equation}
	\partial_t u_A(x,t)
	=
	D_A \Delta u_A(x,t)
	+
	\mathcal{F}_A\!\big(u_A(x,t)\big), 
\end{equation}
\begin{equation}
	\partial_t u_T(x,t)
	=
	D_T \Delta u_T(x,t)
	+
	\mathcal{F}_T\!\big(u_T(x,t),\,u_A(x,t)\big), 
\end{equation}
where $D_A$ and $D_T$ denote diffusion coefficients, $\Delta$ is the spatial Laplacian, $\mathcal{F}_A$ is an autonomous reaction term for amyloid-beta,
and $\mathcal{F}_T$ is a coupled reaction term for tau.
In the coupled setting, the amyloid reaction operator $\mathcal{F}_A$ is fixed from the previously trained autonomous amyloid model, and only the tau reaction operator $\mathcal{F}_T$ is learned in the conditional training stage.
The unidirectional coupling from amyloid-beta to tau is consistent with the widely studied amyloid cascade hypothesis\cite{hardy1992alzheimer}, in which amyloid accumulation precedes and influences subsequent tau pathology.

A key challenge in learning this coupled system is that tau and amyloid-beta
measurements are not paired across subjects or time points.
To address this issue, we compute a population-averaged initial condition
$\bar{u}_A^{(0)}$ for amyloid-beta using the earliest available measurements.
Starting from this population-averaged state, we evolve amyloid-beta forward in time using a trained amyloid-beta model, generating amyloid-beta trajectories.
These generated trajectories are then paired with tau data and used as training inputs for the tau-coupled model.
This strategy enables learning of tau dynamics conditioned on amyloid-beta despite the absence of paired multimodal observations.
This approach assumes that the population-level amyloid trajectory provides a reasonable surrogate for subject-specific amyloid progression when paired multimodal longitudinal observations are unavailable.

The population-averaged initial condition is computed using data from the 111 patients in training the amyloid reaction dynamics. 

\subsection{Left brain}
The following models are trained for 200 epochs using Adam with initial learning rate $2\times 10^{-3}$, halved every 20 epochs.
\begin{table}[H]
	\centering
	\begin{tabular}{|c|c|c|}\hline
		NN architecture& \multicolumn{2}{|c|}{Tanh, $P=4096$}\\\hline
		$[P,M,P]$& $\mathrm{Acc}_1$ & $\mathrm{Acc}_2$ \\\hline
		M = 32  & 81.5\%  &80.1\% \\\hline
		M = 64   &  81.5\% & 80.2\% \\\hline
		M = 128   & 81.7\% & 80.3\% \\\hline
		M = 256 & 81.7\% & 80.4\%\\\hline
		M = 512 & 81.4\% & 80.0\%\\\hline
	\end{tabular}
	\caption{Left hemisphere coupled tau model conditioned on synthetic amyloid trajectories. 
		One-hidden-layer $\tanh$ networks trained with weight decay $10^{-2}$ using $P=4096$ eigenfunctions. 
		The best accuracy achieved is 81.7\% at $M=256$.}
	\label{tab:coupled-left-tau-exp}
\end{table}

\subsection{Right brain}

The following models are trained for 200 epochs using Adam with initial learning rate $2\times 10^{-3}$, halved every 20 epochs.
\begin{table}[H]
	\centering
	\begin{tabular}{|c|c|c|}\hline
		NN architecture & \multicolumn{2}{|c|}{Tanh, $P=4096$}\\\hline
		$[P,M,P]$& $\mathrm{Acc}_1$ & $\mathrm{Acc}_2$ \\\hline
		M = 32  & 81.2\%  & 79.8\% \\\hline
		M = 64   & 81.0\%  & 79.7\%\\\hline
		M = 128  &80.9\%  & 79.6\%\\\hline
		M = 256  &81.3\%  & 79.9\%\\\hline
		M = 512 & 81.1\% & 79.8\% \\\hline
	\end{tabular}
	\caption{Right hemisphere coupled tau model conditioned on synthetic amyloid trajectories. 
		One-hidden-layer $\tanh$ networks trained with weight decay $10^{-2}$ using $P=4096$ eigenfunctions. 
		The best accuracy achieved is 81.3\% at $M=256$.}
	\label{tab:coupled-right-tau-exp}
\end{table}

\section{Derivation of the Optimality System and Forward--Backward Sweep Method}
\label{sec:appendix-derivation-control}

We now derive the adjoint equation by introducing a Lagrange multiplier
$p(t)\in\mathbb{R}^P$ and defining the Lagrangian
\begin{equation}\label{eq:lagrangian_cont_beta}
	\begin{aligned}
		\mathcal{L}(\beta,C,p)
		=
		\int_0^T
		\Big(
		\|\beta(t)\|^2 + \alpha(t) C(t)^2
		\\ 
		+
		p(t)^\top
		\big[
		\dot{\beta}(t)
		+ D\Lambda\beta(t)
		- \mathcal{G}_\theta(\beta(t))
		+ C(t)\beta(t)
		\big]
		\Big)\,dt.    
	\end{aligned}
\end{equation}

Let $\delta\beta$ be an arbitrary variation satisfying $\delta\beta(0)=0$.
Using integration by parts,
\[
\int_0^T p(t)^\top \delta\dot{\beta}(t)\,dt
=
\big[p(t)^\top \delta\beta(t)\big]_{0}^{T}
-
\int_0^T \dot{p}(t)^\top \delta\beta(t)\,dt,
\]
and the Fr\'echet derivative
\[
\delta \mathcal{G}_\theta(\beta)
=
J_\mathcal{G}(\beta)\,\delta\beta,
\]
where $J_\mathcal{G}(\beta)$ denotes the Jacobian of $\mathcal{G}_\theta$,
the first variation of $\mathcal{L}$ with respect to $\beta$ reads
\begin{equation*}
	\begin{aligned}
		\delta_\beta \mathcal{L}
		&= p(T)^\top \delta\beta(T) \\ 
		& + \int_0^T \Big(2\beta - \dot{p}+ D\Lambda^\top p - J_\mathcal{G}(\beta)^\top p
		+
		C p
		\Big)^\top
		\delta\beta
		\,dt.     
	\end{aligned}
\end{equation*}

Since $\delta\beta(t)$ and $\delta\beta(T)$ are arbitrary, stationarity
$\delta_\beta \mathcal{L}=0$ yields the continuous adjoint system
\begin{equation}\label{eq:adjoint_cont_beta}
	\begin{cases}
		-\dot{p}(t)
		=
		\big(
		-D\Lambda
		+
		J_\mathcal{G}(\beta(t))
		-
		C(t)I
		\big)^\top p(t)
		- 2\beta(t),\\[2mm]
		p(T)=0, \qquad t\in(0,T).
	\end{cases}
\end{equation}
Note that $\Lambda^\top=\Lambda$ since $\Lambda$ is diagonal.

Next, we consider variations with respect to the control $C$.
The first variation of $\mathcal{L}$ with respect to $C$ is given by
\[
\delta_C \mathcal{L}
=
\int_0^T
\big(
2\alpha(t) C(t) + p(t)^\top \beta(t)
\big)\,
\delta C(t)\,dt.
\]
Therefore, the first-order optimality condition in the unconstrained case is
\begin{equation}\label{eq:optimality_cont_beta}
	2\alpha(t) C^*(t) + p(t)^\top \beta(t) = 0,
	\qquad
	C^*(t)
	=
	-\frac{p(t)^\top \beta(t)}{2\alpha(t)}.
\end{equation}

When $\mathcal{G}_\theta$ is represented by a neural network, the Jacobian
$J_\mathcal{G}(\beta)$ is not formed explicitly. Instead, the adjoint term
$J_\mathcal{G}(\beta(t))^\top p(t)$ in \eqref{eq:adjoint_cont_beta} is evaluated
via automatic differentiation as follows,
\begin{equation}\label{eq:auto-diff}
	J_\mathcal{G}(\beta)^\top p
	=
	\nabla_\beta \big(p^\top \mathcal{G}_\theta(\beta)\big).    
\end{equation}


The optimal control problem is then solved using a forward--backward sweep method.
Let $t_n=n\Delta t$, $n=0,\dots,N$, $\Delta t=T/N$.

\begin{algorithm}[H]
	\caption{Forward--backward sweep for computing the optimal control}
	\begin{algorithmic}[1]
		\State \textbf{Input:}
		initial state $\beta_0$, parameters $D,\Lambda,\alpha(t)$,
		time step $\Delta t$, final time $T$,
		neural network $\mathcal{G}_\theta$,
		relaxation parameter $\omega\in(0,1]$
		\State \textbf{Initialize:}
		$C_n^{(0)}=0$, $n=0,\dots,N-1$
		\For{$k=0,1,2,\dots$ until convergence}
		\State \textbf{Forward solve (state equation):}
		\State Set $\beta_0^{(k)}=\beta_0$
		\For{$n=0,\dots,N-1$}
		\State Solve the semi-implicit update
		\[
		\big(I+\Delta t(D\Lambda + C_n^{(k)} I)\big)\beta_{n+1}^{(k)}
		=
		\beta_n^{(k)} + \Delta t\, \mathcal{G}_\theta(\beta_n^{(k)}).
		\]
		\EndFor
		\State \textbf{Backward solve (adjoint equation):}
		\State Set $p_N^{(k)}=0$
		\For{$n=N-1,\dots,0$}
		\State Compute the adjoint update
		\begin{equation*}
			\begin{aligned}
				p_n^{(k)} = & \big(I+\Delta t(D\Lambda + C_n^{(k)} I)\big)^\top p_{n+1}^{(k)} \\ 
				& - 2\Delta t\,\beta_n^{(k)}
				+ \Delta t\, J_\mathcal{G}(\beta_n^{(k)})^\top p_{n+1}^{(k)},
			\end{aligned}
		\end{equation*}
		where $J_\mathcal{G}(\beta_n^{(k)})^\top p_{n+1}^{(k)}$ is evaluated as \eqref{eq:auto-diff} by automatic differentiation.
		\EndFor
		\State \textbf{Control update:}
		\For{$n=0,\dots,N-1$}
		\State Compute the pointwise update
		\[
		\tilde C_n^{(k+1)}
		=
		-\frac{(p_{n+1}^{(k)})^\top \beta_{n+1}^{(k)}}{2\alpha_{n+1}}.
		\]
		\State Relaxation:
		\[
		C_n^{(k+1)}
		=
		(1-\omega)C_n^{(k)}+\omega \tilde C_n^{(k+1)}.
		\]
		\EndFor
		\EndFor
		\State \textbf{Output:}
		optimal control $\{C_n\}_{n=0}^{N-1}$ and state trajectory
		$\{\beta_n\}_{n=0}^{N}$
	\end{algorithmic}
\end{algorithm}

We compare this approach with two other approaches for solving the optimal control problem. 
The first one parameterizes the dosing intensity as a function of time using a neural network and optimizes the network parameters by gradient descent. The other one employs a piecewise-constant control on the discrete time grid, and minimizes the
time-discrete objective by gradient descent. Full mathematical details for these two approaches and comparisons are presented in \ref{sec:appendix-other-methods-control}.
The forward-backward sweep method is adopted to compute our experiment results. 

\section{Two other approaches for optimal control}\label{sec:appendix-other-methods-control}

\subsection{Neural network parameterized control} 
We parametrize the control using a shallow neural network
\begin{align}
	\widetilde C_\theta(t)
	&= \sum_{i=1}^{m} c_i\, \sigma(t + b_i),
	\qquad \theta := \{(c_i,b_i)\}_{i=1}^m .
\end{align}

Since the dosing intensity is required to be nonnegative, we introduce a
smooth softplus transformation and define the control function by
\begin{equation}
	C_\theta(t)
	=
	\operatorname{softplus}\big(\widetilde C_\theta(t)\big)
	:=
	\frac{1}{\kappa}\log\!\left(1+e^{\kappa \widetilde C_\theta(t)}\right),
	\quad \kappa = 1 ,
\end{equation}
which guarantees $C_\theta(t)\ge 0$ while preserving differentiability
with respect to the network parameters $\theta$.

The state equation is given by 
\begin{align}
	\begin{cases}
		\partial_t \beta(t) + D \Lambda \beta(t)
		= \mathcal{G}(\beta(t)) - C_\theta(t)\,\beta(t),
		& t\in(0,T), \\[4pt]
		\beta(0) = \beta_0 .
	\end{cases}
\end{align}
The objective functional is 
\begin{align}
	J(C_\theta)
	&=
	\displaystyle
	\int_0^T
	\left(
	\|\beta(t)\|^2
	+
	\alpha(t) \, |C_\theta(t)|^2
	\right)\, dt .
\end{align}
The optimal control problem becomes
\begin{align}
	\min_{\theta}
	\; J(C_\theta)
	\quad \text{subject to the state equation above.}
\end{align}

Let $t_n = n\Delta t$, $n=0,\dots,N$, with $\Delta t = T/N$.  
The state equation is advanced using the semi-implicit scheme
\begin{equation}
	\frac{\beta^{n+1}-\beta^n}{\Delta t}
	+ D\Lambda \beta^{n+1}
	=
	g(\beta^n) - C_\theta(t_n)\,\beta^n ,
\end{equation}
where $C_\theta(t)$ denotes the softplus-transformed neural-network control.

Given the resulting discrete states $\{\beta^n\}_{n=0}^N$, the continuous objective is approximated by the following time-discrete loss function:
\begin{equation}
	\mathcal{L}(\theta)
	=
	\sum_{n=0}^{N-1}
	\Delta t
	\left(
	\|\beta^n\|^2
	+
	\alpha(t_n) |C_\theta(t_n)|^2
	\right).
\end{equation}

\subsection{Discrete-time piecewise-constant control}
We take the control function to be piecewise constant in time,
\[
C(t) \approx C^n, \qquad t\in[t_n,t_{n+1}).
\]
The control function is then approximated by a vector 
\begin{align}
	\mathbf C := (C^0,\dots,C^{N-1}) \in \mathbb{R}^N ,
\end{align}
where $C^n$ denotes the control (dosing intensity) applied on the
time interval $[t_n,t_{n+1})$. 

Since the dosing intensity is required to be nonnegative, we introduce an
unconstrained auxiliary variable 
$\widetilde{\mathbf C}=(\widetilde C^0,\dots,\widetilde C^{N-1})\in\mathbb{R}^N$
and define the discrete control through a smooth softplus transformation
\begin{equation}
	C^n = \operatorname{softplus}(\widetilde C^n)
	:= \frac{1}{\kappa}\log\!\left(1+e^{\kappa \widetilde C^n}\right),
	\quad \kappa = 1,
\end{equation}
for $n=0,\dots,N-1$.
This reparameterization guarantees $C^n\ge 0$ while keeping the optimization problem fully unconstrained and differentiable with respect to $\widetilde{\mathbf C}$.


Let $t_n = n\Delta t$, $n=0,\dots,N$, with $\Delta t = T/N$.
The state equation is advanced using the semi-implicit scheme
\begin{equation}
	\frac{\beta^{n+1}-\beta^n}{\Delta t}
	+ D\Lambda \beta^{n+1}
	=
	\mathcal{G}(\beta^n) - C^n\,\beta^n. 
\end{equation}


Given the discrete states $\{\beta^n\}_{n=0}^N$, we approximate the
continuous objective by the time-discrete loss
\begin{equation}
	\mathcal{L}(\widetilde{\mathbf C})
	=
	\sum_{n=0}^{N-1}
	\Delta t
	\left(
	\|\beta^n\|_{L^2(\Omega)}^2
	+
	\alpha(t_n)|C^n|^2
	\right),
\end{equation}
where $C^n=\operatorname{softplus}(\widetilde C^n)$.

\begin{figure}[H]
	\centering
	\includegraphics[width=\linewidth]{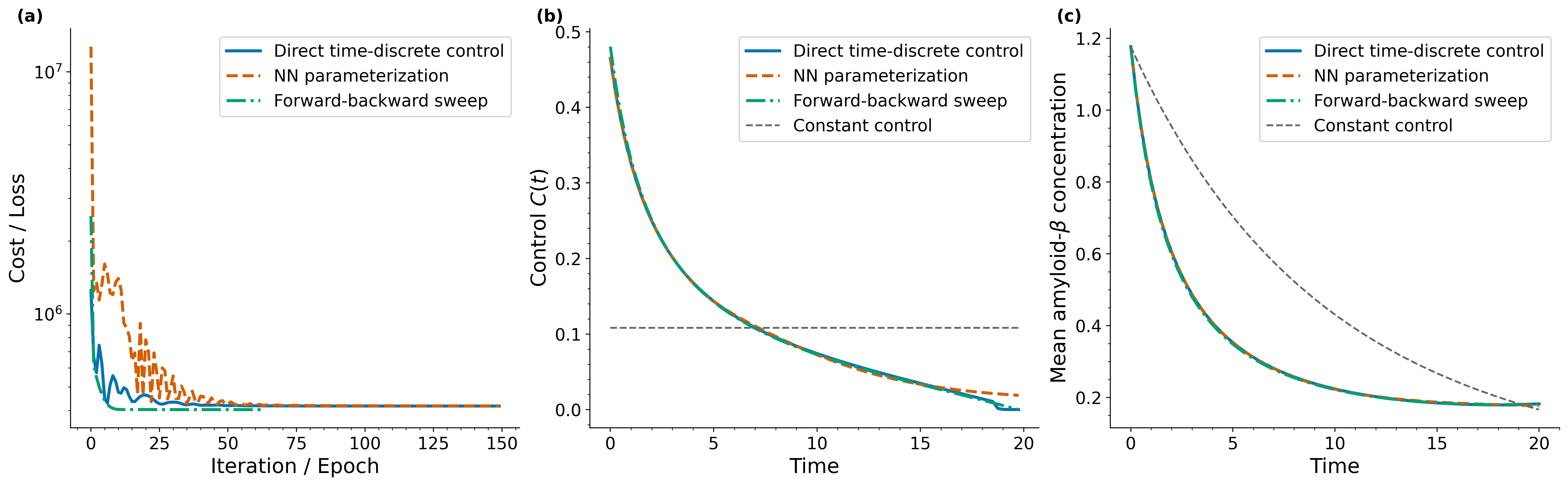}
	\caption{Numerical comparison of the three solvers under constant penalty $\alpha=4\times 10^{5}$ in~\eqref{eq:cost}.
		\textbf{(Left)} objective value vs.\ optimization iteration;
		\textbf{(center)} optimized dosing intensity $C(t)$ vs.\ time;
		\textbf{(right)} spatially averaged amyloid burden vs.\ time.
		Curves correspond to the adjoint-based forward--backward sweep (FBS), the neural-network control parametrization, and the discrete-time piecewise-constant control formulation (Appendix~\ref{sec:appendix-optimal-control}).}
	\label{fig:constant-control-penalty-comparison}
\end{figure}

Figure~\ref{fig:constant-control-penalty-comparison} reports the comparison under constant~$\alpha$ in a single row of three panels: discrete-time objective values along the optimization run (left), optimized dosing trajectories $C(t)$ (center), and spatially averaged amyloid burden under each method (right).
The forward--backward sweep based on the adjoint formulation attains the lowest objective value and the fastest convergence in these panels.

\begin{figure}[H]
	\centering
	\includegraphics[width=\linewidth]{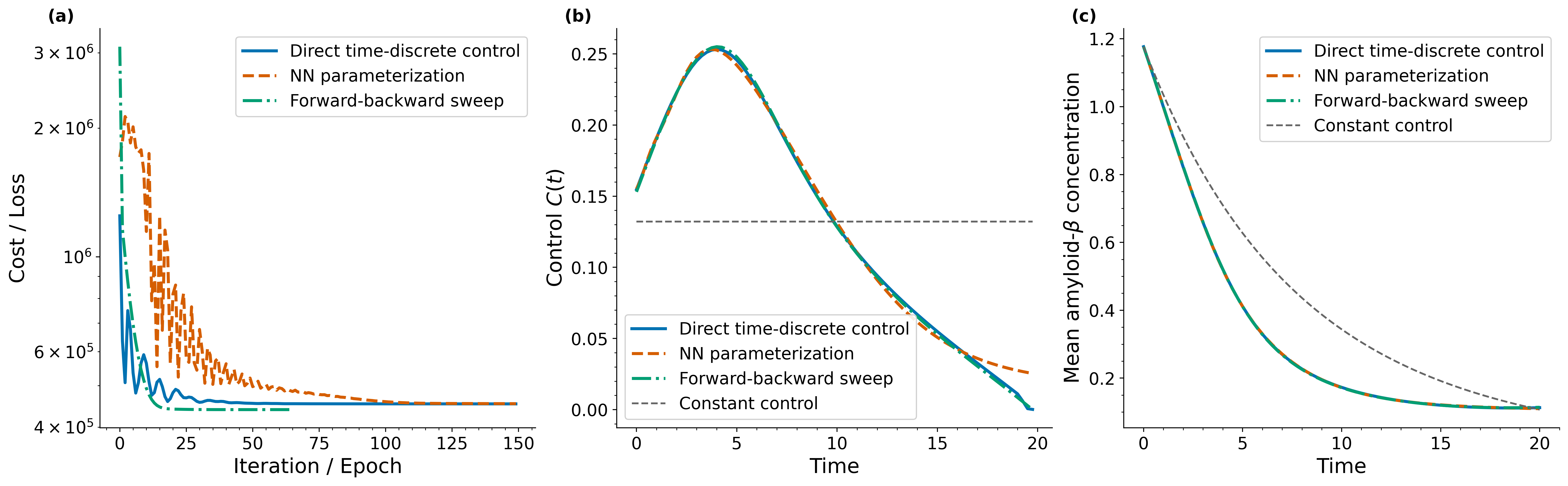}
	\caption{Numerical comparison of the three solvers under the decaying penalty $\alpha(t)=\alpha_1+\alpha_2 e^{-t/\tau}$ with $(\alpha_1,\alpha_2,\tau)=(10^{5},\,1.6\times 10^{6},\,1.5)$.
		\textbf{(Left)} objective value vs.\ optimization iteration;
		\textbf{(center)} optimized dosing intensity $C(t)$ vs.\ time;
		\textbf{(right)} spatially averaged amyloid burden vs.\ time.
		Curve styles match Figure~\ref{fig:constant-control-penalty-comparison}.}
	\label{fig:decaying-control-penalty-comparison}
\end{figure}

We next repeat the comparison using a time-dependent penalty $\alpha(t)$ that decays over time, as used in the generalized objective in Appendix~\ref{sec:appendix-optimal-control},
\begin{equation}\label{eq:decaying-alpha-penalty}
	\alpha(t) = \alpha_1 + \alpha_2 e^{-t / \tau}. 
\end{equation}
In our experiment, we set $\alpha_1 = 100000$, $\alpha_2 = 1600000$ and $\tau = 1.5$. 
Figure~\ref{fig:decaying-control-penalty-comparison} shows the same three-panel layout for the decaying penalty~\eqref{eq:decaying-alpha-penalty}.
The forward--backward sweep again achieves the lowest loss and the fastest convergence.

\section{Implementation details of VR}\label{app:vr-details}
The virtual reality (VR) platform that visualizes the digital twin (DT) environment is engineered using the Unity 3D engine (v. 2022.3.61f1) and deployed on the Meta Quest 3 spatial computing headset.  To anchor the user's spatial cognition, a full-scale anatomical human avatar is integrated as a contextual spatial reference. To enable the unoccluded observation of internal biomarkers within the human shell, we engineered a dynamic material-swapping rendering pipeline. The avatar's brain mesh utilizes a multi-state Physically Based Rendering (PBR) system:
\begin{itemize}
	\item Brain Shell: Under the default observation mode, the brain mesh is rendered with an opaque material. Upon triggering the animation mode, the system dynamically replaces the native material with a custom alpha-blended transparent shader (e.g., modulating the alpha channel to $\alpha \approx 0.2-0.4$). This creates a semi-transparent ``glass envelope" effect, minimizing visual occlusion.
	\item Biomarker Mapping \& Rendering: Utilizing a custom shader-based visualizer framework, simulated spatiotemporal data of tau protein and amyloid-$\beta$ concentrations are mapped onto high-resolution 3D mesh. By translating discrete temporal simulation frames into continuous color gradients on the organ surface, this approach allows for precise, three-dimensional observation of biomarker dynamics across the simulated disease trajectory.
\end{itemize}
Furthermore, the system introduces a multimodal 6-Degree-of-Freedom (6-DoF) interaction paradigm, effectively bridging the gap between abstract computational simulations and tactile spatial exploration:
\begin{itemize}
	\item Direct Spatial Manipulation: The virtual brain is treated as a manipulable physical entity. Leveraging rigid-body kinematics and grab interactors from the Meta XR All-in-One SDK, users can directly grasp, rotate, and inspect the complex 3D geometry in real-time. To maintain spatial stability and prevent tracking drift after user interaction, a kinematic state-restoration algorithm is employed to zero physical inertia (i.e., linear and angular velocities) and strictly reset local transforms during visualization resets.
	\item Temporal Scrubbing and Heads-Up Display (HUD): A dynamically tracked world-space HUD provides a high-level temporal control interface. By implementing a Pointable Canvas framework, we bridged traditional 2D Unity UI components, such as the disease timeline slider, with VR-native ray-casting and direct poke interactions. This allows users to intuitively ``scrub" through the temporal progression. An event-driven synchronization architecture ensures that each slider increment triggers a frame-locked update across the 3D biomarker gradients, the numerical age readout, and the global simulation index.
\end{itemize}
By providing immediate and immersive visual feedback on the brain surface, this system significantly improves the interpretability of complex, data-driven predictions. This integration highlights the immense potential of combining temporally dynamic DT with spatial computing to accelerate hypothesis generation and foster translational communication between computational researchers and
clinicians.

\showacknow{} 

\bibsplit[3]

\bibliography{pnas-sample}

\end{document}